\documentclass[letterpaper]{article} 
\usepackage{aaai25}  
\usepackage{times}  
\usepackage{helvet}  
\usepackage{courier}  
\usepackage[hyphens]{url}  
\usepackage{graphicx} 
\usepackage{natbib}  
\usepackage{caption} 
\usepackage{algorithm}
\usepackage{algpseudocode}

\usepackage{newfloat}
\usepackage{listings}
\DeclareCaptionStyle{ruled}{labelfont=normalfont,labelsep=colon,strut=off} 
\floatstyle{ruled}
\newfloat{listing}{tb}{lst}{}
\floatname{listing}{Listing}
\usepackage{amsmath,amsthm,amssymb}

\usepackage{tabularx}
\usepackage{multirow}
\usepackage{makecell}

\usepackage{booktabs}  
\usepackage{rotating}
\usepackage{pdflscape}
\usepackage{cancel}
\usepackage{enumerate, cases}
\usepackage{thmtools,thm-restate}
\usepackage{mathtools}
\usepackage[export]{adjustbox}
\usepackage{pifont}
\usepackage{array}
\usepackage{enumitem}
\setlist[itemize]{noitemsep, nolistsep}

\renewcommand{\phi}{\varphi}
\renewcommand{\epsilon}{\varepsilon}

\newcommand{\el}{\end{flushleft}}
\newcommand{\bl}{\begin{flushleft}}

\newcommand{\bR}{\mathbb{R}}

\newcommand{\cA}{\mathcal{A}}

\newcommand{\cC}{\mathcal{C}}

\newcommand{\cH}{\mathcal{H}}

\newcommand{\cL}{\mathcal{L}}
\newcommand{\cM}{\mathcal{M}}

\newcommand{\mat}[1]{\mathbf{#1}}

\theoremstyle{plain}

\newtheorem{observation}{Observation}

\newtheorem{theorem}{Theorem}[section]
\theoremstyle{remark}
\newtheorem{remark}[theorem]{Remark}

\title{On Newton's Method to Unlearn Neural Networks}
\author {
    Bui Thi Cam Nhung*\textsuperscript{\rm 1},
    Xinyang Lu*\textsuperscript{\rm 12},
    Rachael Hwee Ling Sim\textsuperscript{\rm 1},
    See-Kiong Ng\textsuperscript{\rm 12},
    Bryan Kian Hsiang Low\textsuperscript{1}
}
\affiliations {
    \textsuperscript{\rm 1}Department of Computer Science, National University of Singapore\\
    \textsuperscript{\rm 2}Institute of Data Science, National University of Singapore\\
    btcnhung@comp.nus.edu.sg, 
    xinyang.lu@u.nus.edu, 
    rachael.sim@u.nus.edu,
    seekiong@nus.edu.sg, 
    lowkh@comp.nus.edu.sg
}

\begin{document}

\maketitle

\begin{abstract}
    With the widespread applications of \emph{neural networks} (NNs) trained on personal data, machine unlearning has become increasingly important for enabling individuals to exercise their personal data ownership, particularly the ``right to be forgotten'' from trained NNs.
    Since retraining is computationally expensive, we seek approximate unlearning algorithms for NNs that return identical models to the retrained oracle.
    While Newton's method has been successfully used to approximately unlearn linear models, we observe that adapting it for NN is challenging due to degenerate Hessians that make computing Newton's update impossible. 
    Additionally, we show that when coupled with popular techniques to resolve the degeneracy, Newton's method often incurs offensively large norm updates and empirically degrades model performance post-unlearning.
    To address these challenges, we propose CureNewton's method, a principle approach that leverages cubic regularization to handle the Hessian degeneracy effectively.
    The added regularizer eliminates the need for manual finetuning and affords a natural interpretation within the unlearning context.
    Experiments across different models and datasets show that our method can achieve competitive unlearning performance to the state-of-the-art algorithm in practical unlearning settings,  while being theoretically justified and efficient in running time.

\end{abstract}

%

\section{Introduction}
\label{sec:introduction}
    In recent years, there have been more large \emph{machine learning} (ML) models trained on vast amounts of data, including personal data from individual users, for computer vision, natural language processing, and speech recognition applications~\citep{radford23:whisper, achiam23:gpt-4, edpb24:facial-recognition-airport}.
    At the same time, new regulations, such as the General Data Protection Regulation (\citeauthor{gdpr16}) 2016 and the California Consumer Privacy Act (\citeauthor{ccpa18}) 2018, have granted users the ``right to be forgotten'', allowing them to request the timely erasure of their data from ML models.
    In response, the field of \emph{machine unlearning}~\citep{bourtoule21:sisa-unlearning,cao15:towards-forget-summation,nguyen22:survey-unlearn} has emerged and seeks to effectively and efficiently remove the influence of erased training data points from the ML model while preserving model performance on the retained data points.
    
    While \textit{exact unlearning} can be achieved by retraining the ML model with only the retained data points (i.e., excluding those to be erased) from scratch, retraining is computationally expensive and impractical for big datasets and deep \emph{neural networks} (NN). 
    Moreover, the extensive use of non-linear operations and iterative optimization methods in NNs make it difficult to trace the influence of the erased data points on the model parameters and design exact unlearning algorithms for NNs without retraining.\footnote{In contrast, the influence of data points in many non-NN models, such as support vector machines~\citep{cauwenberghs00:incremental-svm}, k-means~\cite{ginart19:making-forget}, random forests~\citep{brophy21:random-forest-unlearn}, can be exactly removed efficiently.}
    Thus, \citet{ginart19:making-forget, guo20:certified-removal} propose \emph{approximate unlearning} algorithms that can efficiently update a trained ML model (at a lower cost than retraining) such that the distribution of the ML models after unlearning is similar to the distribution of models retrained from scratch.\footnote{The definition in \citet{guo20:certified-removal} assumes the learning and unlearning algorithms are randomized algorithms. 
    }
    Specifically, \citet{neel21:descent-to-delete} proposes a gradient-based approximate unlearning algorithm that performs several steps of gradient descent on the retained data points.
    \citet{warnecke2021machine} proposes to estimate and revert the influence of the erased data points on the model parameters using the influence function~\citep{koh2017understanding}.
    The closest work to ours is the second-order approximate unlearning algorithm proposed in \citet{guo20:certified-removal}, which performs few-shot Newton's updates on the retained data points to unlearn linear models with convex losses.

    Second-order approximate unlearning algorithms perform a second-order Taylor expansion of the loss function on the retained data points at the original model parameters (prior to unlearning) to approximate the loss at other parameters, such as the post-unlearning parameters. 
    This approximation, which is a {\it surrogate function} to the loss function on the retained data points, is then minimized using Newton's method to obtain the best parameters for the unlearned model~\cite{guo20:certified-removal}. 
    However, we observe that the Hessian matrix (second-order partial derivatives) of the loss function is often degenerate and non-invertible for NNs, making computing Newton's update impossible (Sec.~\ref{sec:hessian_degeneracy}).\footnote{\citet{golatkar20:selective-forget,guo20:certified-removal} have only proven approximate unlearning of \emph{linear} models with \emph{quadratic/convex losses} where the \emph{Hessians} are often \emph{positive definite}.} 
    In addition, we show that popular techniques to resolve the degeneracy, such as taking the pseudoinverse and adding a damping factor to the Hessian diagonal, often face issues --- they lead to large $\ell_2$-norm Newton's updates and empirically destroy the model performance after unlearning without sufficient hyperparameter tuning (Sec.~\ref{sec:baselines_degeneracy}).

    
    This work proposes a better principled method to tackle Hessian degeneracy, which enables stable computation of Newton's update to unlearn trained NNs. 
    Inspired by the seminal work of \citet{nestero06:cubic-newton}, we propose \emph{\underline{Cu}bic-\underline{re}gularized \underline{Newton}'s} (CureNewton's) method which introduces a cubic regularizer term in the surrogate function capturing the loss on the retained data points (Sec.~\ref{sec:method}). The additional term ensures that our surrogate is an upper bound on the true loss and decides the damping factor without the need for manual fine-tuning.
    Interestingly, the optimized damping factor offers a natural interpretation as the $\ell_2$-distance of the model parameters before and after unlearning.
    As it can automatically adapt to different unlearning requests under the assumption of an $L$-Lipschitz continuous Hessian, CureNewton is applicable in many practical settings such as batch unlearning and sequential unlearning of random training points and classes.
    Empirical evaluation on real-world datasets reveals that CureNewton can effectively and efficiently unlearn a variety of models with competitive performance to state-of-the-art algorithms. Additionally, it is theoretically justified and efficient in terms of running time (Sec.~\ref{sec:exp}).

\section{Related Works}
\label{sec:related_works}
    
    {\bf Exact vs.~approximate unlearning.}
    Ideally, \emph{exact unlearning} should generate a model that is identical to a model retrained from scratch without the requested/erased training data points. Earlier works have proposed meticulous learning frameworks capable of exact unlearning in statistical models such as support vector machines, $k$-means clustering, random forests, and item-based collaborative filtering \citep{brophy21:random-forest-unlearn, cao15:towards-forget-summation, cauwenberghs00:incremental-svm, ginart19:making-forget}. To facilitate exact unlearning in neural networks (NN), the exemplary work of~\citet{bourtoule21:sisa-unlearning} proposes a data-centric method that maintains disjoint data subsets and their corresponding models to reduce the cost of retraining. However, a simple slow-down attack with uniformly distributed requests can significantly increase its cost~\citep{yan22:arcane-exact}. Since precisely unlearning an NN is difficult and the cost of retraining an NN is typically high, recent literature has adopted a notion of {\it approximate unlearning} that seeks an unlearned model similar to the retrained model \citep{ginart19:making-forget, guo20:certified-removal}. This gives rise to various unlearning heuristics, including fine-tuning on random labels, moving in the opposite direction of optimization, or teaching the model to produce unseen-like predictions for the removed points~\citep{chundawat23:bad-teaching,tarun23:fast-unlearn, warnecke2021machine}. However, they often lack guarantees for the approximation to the retrained model to certify removal~\citep{guo20:certified-removal} and measure the amount of information leakage on the set that has been removed~\citep{golatkar20:selective-forget}. Within this taxonomy, our method falls into the category of approximate unlearning.
    
    \vspace{1mm}
    {\bf Weak vs.~strong unlearning.}
    Approximate unlearning can be further categorized into two subclasses based on the granularity level of indistinguishability to the retrained model~\citep{xu23:unlearn-survey}. {\it Weak unlearning} seeks an unlearned model with similar predictions to the retrained model, i.e., indistinguishability in the output space. Such techniques include linear filtration in the logit-based classifiers~\citep{baumhauer22:linear-filtration} and linearizing the final activations via the Neural Tangent Kernel~\citep{golatkar20:forget-ntk}. On the other hand, {\it strong unlearning} methods modify model parameters to obtain indistinguishability in the parameter space. This goal is reflected in the notion of $(\epsilon, \delta)$-certified removal~\citep{guo20:certified-removal}, which is inspired by the definition of differential privacy~\citep{dwork06:differential-privacy}. To tackle strong unlearning, earlier works have used popular optimization techniques to achieve minimal loss on the retained set, thereby resembling retrained models. First-order methods, such as gradient descent and noisy gradient descent, are used in the work of~\citet{neel21:descent-to-delete} and~\citet{ullah2021machine} to unlearn models with a strong assumption of model stability. More effectively, the second-order methods of~\citet{golatkar20:selective-forget, guo20:certified-removal} provide closed-form unlearning updates with lower approximation error to the retrained solution.
    Our work generalizes~\citet{guo20:certified-removal} solution to work for trained NNs, including those with degenerate Hessians.

\section{Preliminaries}
\label{sec:preliminaries}

\subsection{Problem Setting}
\label{sec:problem}

    Let $\cA$ denote a learning algorithm that takes a training set $D$ and returns a model $h \in \mathcal{H}$ by training on $D$. The training set $D$ can be partitioned into two disjoint subsets, $D_e$ of erased data points and $D_r$ of retained data points, i.e., $D_e \cup D_r = D$ and $D_e \cap D_r = \emptyset$. We use $n, n_e, n_r$ to denote the size of $D, D_e, D_r$, respectively. Our goal is to design an unlearning algorithm $\cM$ that takes a tuple $(\cA(D), D, D_e)$ and returns a new model $h_r \in \cH$ that is free of $D_e$'s influence and performs well on $D_r$.

    An unlearning algorithm is deemed exact if $\cM(\cA(D), D, D_e) = \cA(D \setminus D_e)$. Retraining from scratch, i.e. $\cM(\cA(D), D, D_e) = \cA(D_r)$, is an unlearning algorithm that trivially achieves exact unlearning (because $D_r = D \setminus D_e$), but scales poorly with big datasets and deep NNs. Instead of exact unlearning, we aim for an approximate unlearning algorithm that returns an unlearned model identical to the retrained one\footnote{We use this definition instead of {\it distributional indistinguishability} in \cite{guo20:certified-removal} as we will consider models optimized with empirical risk minimization, which are unique.}:
    \begin{equation*}
        \cM(\cA(D), D, D_e) \approx \cA(D_r)\ .
    \end{equation*}
    
    
    

    \subsection{Newton's Method for Unlearning}
    \label{sec:newton}

    Let $\cA(D)$ be the minimizer of the following empirical risk of the model parameterized by $\mat{w} \in \bR^d$ and measured on the training dataset $D$:
    \begin{equation*}
    \textstyle
        \cL_{D}(\mat{w}) = \frac{1}{n} \sum_{\mat{x} \in D} \ell(\mat{x}; \mat{w}) + \cC(\mat{w})\ ,
    \end{equation*}
    where $\ell$ is often a non-convex loss and $\cC$ is a convex regularizer. Let $\mat{w}^* = \cA(D) = \mathop{\arg\min}_{\mat{w} \in \bR^d} \mathcal{L}_D(\mat{w})$ be the parameters of the original model (before unlearning) and $\mat{w}_r^* = \cA(D_r) = \mathop{\arg\min}_{\mat{w} \in \bR^d} \cL_{D_r}(\mat{w})$ be those of the retrained model on the retained data points. To perform unlearning on the original model, we take a second-order Taylor expansion of $\cL_{D_r}$ in the vicinity of $\mat{w}^*$:
    \begin{equation}
        \label{eqn:quadratic_approximation}
        \begin{aligned}
            \widetilde{\cL}_{D_r}(\mat{w}) = &\quad \cL_{D_r}(\mat{w}^*) + \langle \mat{g}^r_{\mat{w}^*}, \mat{w} - \mat{w}^* \rangle \\
        &{\textstyle + \frac{1}{2} \left\langle \mat{H}^r_{\mat{w}^*} (\mat{w} - \mat{w}^*), \mat{w} - \mat{w}^* \right\rangle\ ,}
        \end{aligned}
    \end{equation}
    where $\mat{g}^r_{\mat{w}^*} := \nabla \cL_{D_r}(\mat{w}^*)$ and $\mat{H}^r_{\mat{w}^*} := \nabla^2 \cL_{D_r}(\mat{w}^*)$. We refer to $\widetilde{\cL}_{D_r}$ as the {\it surrogate function} to $\cL_{D_r}$.
    Subsequently, we aim to minimize $\widetilde{\cL}_{D_r}(\mat{w})$ to obtain the unlearned model parameters that are close to the retrained parameters $\mat{w}_r^*$.
    By solving for the first-order optimality condition of the local quadratic approximation in Eq.~\ref{eqn:quadratic_approximation}, i.e. $\nabla \widetilde{\cL}_{D_r}(\mat{w}) = \mat{0}$, we obtain the closed-form parameters update:
    \begin{align}
    \mat{w} 
    &= \mat{w}^* - (\mat{H}^r_{\mat{w}^*})^{-1} \mat{g}^r_{\mat{w}^*} \label{eqn:newton_unlearn_update} \\
    \textstyle
    &= {\textstyle \mat{w}^* - (\mat{H}_{\mat{w}^*} - \frac{n_e}{n} \cdot \mat{H}^e_{\mat{w}^*})^{-1} (\mat{g}_{\mat{w}^*} - \frac{n_e}{n} \cdot \mat{g}^e_{\mat{w}^*})\ .} \label{eqn:newton_unlearn_update_unpack}
    \end{align}
    Eq.~\ref{eqn:newton_unlearn_update} constitutes a Newton-like update for unlearning (through second-order optimization on $D_r$) and Eq.~\ref{eqn:newton_unlearn_update_unpack} follows from the linearity of differentiation, which elaborates the connection of the unlearned parameters to the erased data (through the gradient $g_{\mat{w}^*}^e$ and $\mat{H}_{\mat{w}^*}^e$ measured on $D_e$).
    If the original model converges to the first-order stationary point, Eq.~\ref{eqn:newton_unlearn_update_unpack} can be simplified with $\mat{g}_{\mat{w}^*} = \mat{0}$.
    \citet{guo20:certified-removal} proposes to use Newton's method (Eq.~\ref{eqn:newton_unlearn_update}) for unlearning ML models with quadratic/convex losses, where the Hessians are often positive definite (p.d.) and $\mat{w}$ is the global minimizer of the surrogate function $\widetilde{\cL}_{D_r}(\mat{w})$.

    \section{Understanding Newton's Method for Unlearning Trained NNs}
    \label{sec:motivation}

    \subsection{Problems with Degenerate Hessians}
    \label{sec:hessian_degeneracy}

    \begin{figure}
        \centering
        \includegraphics[width=0.9\linewidth]{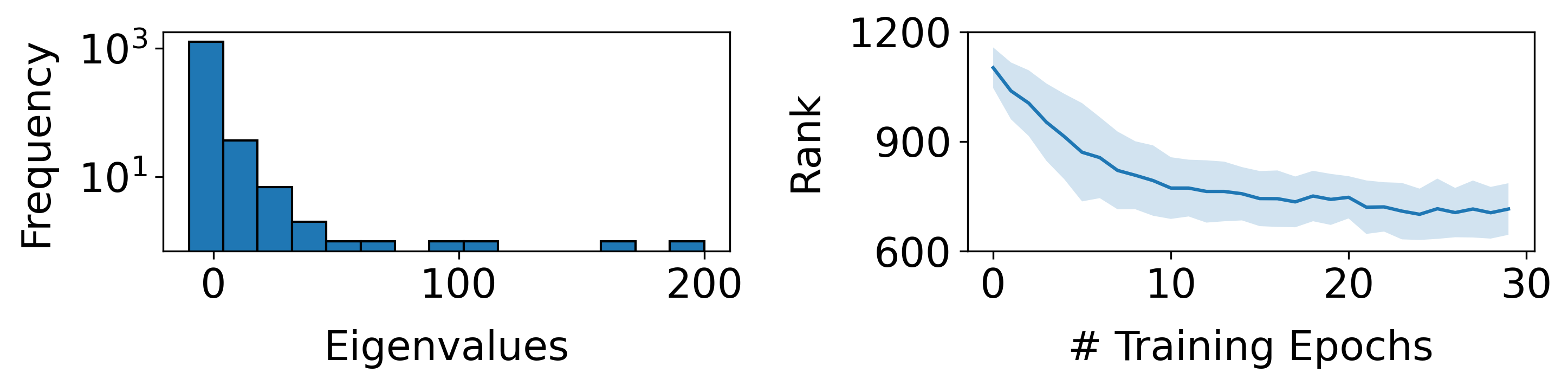}
        \caption{The eigenspectrum (left) and the training rank dynamics (right) of $\mat{H}_{\mat{w}^*}^r$ in a 2-layer CNN trained on the FashionMNIST dataset (for $3$ random runs).
        }
        \label{fig:hessian}
        \vspace{-3mm}
    \end{figure}
    
    To adopt Newton's method for unlearning trained NNs, we assume that $\mat{H}_{\mat{w}^*}^r$ in Eq.~\ref{eqn:newton_unlearn_update} is non-degenerate and invertible. However, this assumption often fails to hold in reality. Previous works have found that for many trained NNs, the eigenspectrum of the Hessians often encompasses a large number of zero eigenvalues with a fast decaying tail~\citep{sagun17:empirical-analysis-hessian, pennington17:geometry-loss, ghorbani19:investigation-hessian}, as also seen in Fig.~\ref{fig:hessian} (left). To explain this phenomenon, theoretical bounds have tied the rank deficiency (which implies the increasing degree of degeneracy) of the Hessians to the effective number of model parameters that naturally diminishes during training~\citep{singh21:rank-hessian, singh23:hessian-cnn}, as also seen in Fig.~\ref{fig:hessian} (right). Generally, these analyses suggest that the Hessians are often severely degenerate as the nature of convergence and thus are non-invertible in Newton's method for unlearning trained NNs. Consequently, we are motivated to answer the question: {\bf How can we adopt Newton's method for unlearning trained NNs in the face of ubiquitous degenerate Hessians (Observation~\ref{obs:hessian_eigenspectrum})?}
    
    
    \begin{observation}
        \label{obs:hessian_eigenspectrum}
        At convergence, $\mat{H}_{\mat{w}^*}^r \in \bR^{d \times d}$ is degenerate whose eigenvalues (in non-increasing order) are $\{\lambda_1, \ldots, \lambda_k, 0, \ldots, 0\}$ ($k < d$) and the corresponding eigenvectors are $\{\mat{v}_1, ..., \mat{v}_d\}$.
    \end{observation}

    \subsection{Sensitivity of Degeneracy Baselines}
    \label{sec:baselines_degeneracy}
    
    Before properly tackling the degenerate Hessians, we first discuss the cons of two baselines: (i) taking the pseudoinverse of the Hessian matrix and (ii) damping the degenerate Hessian with a small diagonal matrix.

    {\bf Pseudoinverse} is a general method to calculate the inverse of any matrix that can be non-invertible. Given $\mat{A} \in \bR^{m \times n}$ whose singular value decomposition is $\mat{A} = \mat{U} \boldsymbol{\Sigma} \mat{V}^T$ where $\mat{U} \in \bR^{m \times m}$, $\mat{V} \in \bR^{n \times n}$ are orthogonal matrices and $\boldsymbol{\Sigma} \in \bR^{m \times n}$ is a diagonal matrix of singular values, then $\mat{A}^\dagger = \mat{V} \boldsymbol{\Sigma}^\dagger \mat{U}^T$ is the pseudoinverse of $\mat{A}$ where $\boldsymbol{\Sigma}^\dagger \in \bR^{n \times m}$ is constructed by taking the reciprocal of $\boldsymbol{\Sigma}$'s non-zero singular values.
    
    \begin{remark}
    \label{rmk:pinv}
        Taking the pseudoinverse is equivalent to find $\Delta = \mat{w}^* - \mat{w}$ with the smallest norm to the least squares problem $\min_{\Delta} \lVert \mat{H}_{\mat{w}^*}^r \Delta - \mat{g}_{\mat{w}^*}^r \rVert$. Under Obs.~\ref{obs:hessian_eigenspectrum}, the unique solution has norm $\lVert \Delta \rVert = \sqrt{\sum_{i=1}^k \frac{1}{\lambda_i^2} | \mat{v}_i^\top \mat{g}_{\mat{w}^*}^r |^2}$, which is the minimal norm among others (see derivation in Appendix~\ref{sec:proof_rmk_pinv}).
        Despite that, as the eigenspectrum of $\mat{H}_{\mat{w}^*}^r$ contains many near-zero eigenvalues (besides the exact-zero eigenvalues), taking the inverse of the small eigenvalues may lead to offensively large norm updates.
    \end{remark}

    {\bf Damping} adds a small positive diagonal matrix $\gamma \mat{I}$ to regularize a square matrix $\mat{A} \in \bR^{n \times n}$, where $\mat{I} \in \bR^{n \times n}$ is the identity matrix and $\gamma \in \bR^{+}$ is a damping factor. Consequently, the eigenvalues of $\mat{A} + \gamma \mat{I}$ are uniformly shifted by $\gamma$, while its eigenvectors remain unchanged.
    
    \begin{remark}
        \label{rmk:damping}
             Damping finds the solution to the regularized least squares problem $\min_{\Delta} \lVert \mat{H}_{\mat{w}^*}^r \Delta - \mat{g}_{\mat{w}^*}^r \rVert^2 + \gamma \lVert \Delta \rVert^2$. Under Obs.~\ref{obs:hessian_eigenspectrum}, the unique solution has norm $\lVert \Delta \rVert = \sqrt{\sum_{i=1}^k \frac{1}{(\lambda_i + \gamma)^2} | \mat{v}_i^\top \mat{g}_{\mat{w}^*}^r |^2 + \sum_{j=k+1}^d \frac{1}{\gamma^2} | \mat{v}_j^\top \mat{g}_{\mat{w}^*}^r |^2}$ (see derivation in Appendix~\ref{sec:proof_rmk_damping}).
        As $\gamma \rightarrow 0$, we recover the original Newton's update. Nevertheless, taking the inverse of infinitesimal $\gamma$ may lead to offensively large norm updates.
    \end{remark}

    Our experiments affirm the sensitivity of both baselines to the degenerate Hessian matrix of a converged model (see Sec.~\ref{sec:exp:batch_unlearning}).
    Based on our analysis, especially of the damping method, we put forward a question: {\bf Is it possible to optimize for $\gamma$ that helps Newton's method do well in unlearning?} The next section is our answer to this question.
    
    \section{\underline{Cu}bic-\underline{re}gularized \underline{Newton}'s (CureNewton's) Method for Unlearning}
    \label{sec:method}

    \subsection{Regularized Surrogate}
    \label{sec:reg_approximation}
    We will introduce a cubic regularizer to the surrogate function $\widetilde{\cL}_{D_r}$ which is beneficial for us to find the optimal $\gamma$ in damped Newton's method for unlearning, as we will show later.
    To begin, let us assume $\mat{H}_{\mat{w}}^r$ is continuous on $\bR^d$ with Lipschitz constant $2L$ ($L > 0$):
    \begin{equation*}
        \lVert \mat{H}_{\mat{w}_1}^r - \mat{H}_{\mat{w}_2}^r \rVert \leq 2L \lVert \mat{w}_1 - \mat{w}_2 \rVert, \quad \forall \mat{w}_1, \mat{w}_2 \in \bR^d\ . 
    \end{equation*}
    Following the work of~\citet{nestero06:cubic-newton}, we add a cubic term to the local quadratic approximation of $\cL_{D_r}$:
    \begin{equation}
        \label{eq:cubic_approx_true}
        \begin{aligned}
            \widetilde{\cL}_{D_r}(\mat{w}) &= 
            \cL_{D_r}(\mat{w}^*) + \left\langle \mat{g}^r_{\mat{w}^*}, \mat{w} - \mat{w}^* \right\rangle \\
            &+ \frac{1}{2} \left\langle \mat{H}^r_{\mat{w}^*} (\mat{w} - \mat{w}^*), \mat{w} - \mat{w}^* \right\rangle + \frac{L}{3} \lVert \mat{w} - \mat{w}^* \rVert^3 \ .
        \end{aligned}
    \end{equation}
    In a similar endeavor to find the minimizer of the non-regularized surrogate function (Eq.~\ref{eqn:quadratic_approximation}), one may be tempted to solve for $\nabla \widetilde{\cL}_{D_r}(\mat{w}) = \mat{0}$ and obtain the following result:
    \begin{equation*}
    \mat{w} = \mat{w}^* - \left( \mat{H}^r_{\mat{w}^*} + L\lVert \mat{w} - \mat{w}^* \rVert \mathbf{I} \right)^{-1} \mat{g}^r_{w^*}\ .
    \end{equation*}
    This is not a closed-form update for $\mat{w}$ because both sides of the equation contain $\mat{w}$. 
    Nevertheless, we can transform it into the damped Newton's update by setting $\gamma := L \lVert \mat{w} - \mat{w}^* \rVert > 0$.
    In the next subsection, we will discuss how to define and optimize $\lVert \mat{w} - \mat{w}^* \rVert$ for the update.
    
    
    \subsection{CureNewton's Method for Unlearning}
    \label{sec:cr_newton}

    Let $\Delta_{\alpha} := -(\mat{H}_{\mat{w}^*}^r + \alpha L \mat{I})^{-1} \mat{g}_{\mat{w}^*}^r$ where $\alpha$ is the damping factor for the Hessian matrix we aim to optimize for. Then, $\Delta_{\alpha} = \mat{w} - \mat{w}^*$ and
    finding the minimizer of the regularized surrogate function (Eq.~\ref{eq:cubic_approx_true}) can be stated as the following optimization problem ($\mathcal{L}_{D_r}(\mat{w}^*)$ is a constant, hence omitted):
    %
    \begin{equation*}
    \resizebox{\linewidth}{!}{
    \label{eq:cubic_primal_problem}
        $
        \min_{\Delta_\alpha \in \bR^d} v_u(\Delta_\alpha) := \left\langle \mat{g}_{\mat{w}^*}^r, \Delta_\alpha \right\rangle + \frac{1}{2} \left\langle \mat{H}_{\mat{w}^*}^r \Delta_\alpha, \Delta_\alpha \right\rangle + \frac{L}{3} \lVert \Delta_\alpha \rVert^3 
        $
    }
    \end{equation*}
    Solving the above problem is difficult because of the non-convexity induced by the cubic term. 
    Despite that, Theorem 10 of~\citet{nestero06:cubic-newton} has shown that we can form the following dual problem of $\min_{\Delta_\alpha \in \bR^d} v_u(\Delta_\alpha)$ with a strong duality\footnote{The duality gap is $\frac{2}{3L} \cdot \frac{\alpha + 2 \lVert \Delta_\alpha \rVert}{(\alpha + \lVert \Delta_\alpha \rVert)^2}  \cdot v_l'(\alpha)^2$.
    This implies a strong duality, i.e. $\min_{\Delta_\alpha \in \bR^d} v_u(\Delta_\alpha) = \sup_{\alpha \in Q} v_l(\alpha)$ if $v_l'(\alpha) = 0$.
    }:
    \begin{equation*}
    \begin{aligned}
        & \sup_{\alpha} & & v_l(\alpha) := -\frac{1}{2} \left\langle \left( \mat{H}_{\mat{w}^*}^r + \alpha L \mat{I} \right)^{-1} \mat{g}_{\mat{w}^*}^r, \mat{g}_{\mat{w}^*}^r \right\rangle - \frac{L}{6} \alpha^3 \\
        & \text{s.t.} & & \alpha \in Q = \{\alpha \in \bR: \mat{H}_{\mat{w}^*}^r + \alpha L \mathbf{I} \succ 0, \alpha \geq 0 \}
    \end{aligned}
    \end{equation*}
    Fortunately, $Q$ is a convex set and $\sup_{\alpha \in Q} v_l(\alpha)$ is a convex constrained optimization problem. 
    Therefore, we can find the optimal $\alpha$ through well-established optimization techniques such as trust-region methods~\citep{jorge06:numerical-optimization}, which is also detailed in Appendix~\ref{app:cure_newton}. In most cases, $\alpha$ will be iteratively optimized by taking a Newton's step:
    %
    \begin{equation*}
    \label{eqn:alpha_update}
    \textstyle
         \alpha_n = \alpha_{n-1} + \frac{v_l'(\alpha_{n-1})}{v_l''(\alpha_{n-1})}\ ,
    \end{equation*}
    where $\alpha = \frac{\gamma}{L}$ and $\gamma$ is initialized by the perturbed smallest eigenvalue of $\mat{H}_{\mat{w}^*}^r$ (so that the Hessian is p.d.) and stop when $|v_l'(\alpha)| = \frac{L}{2} | \lVert \Delta_\alpha \rVert^2 - \alpha^2 | \leq \epsilon$. The optimal $\alpha$ completes our CureNewton's update for unlearning: 
    \begin{equation}
    \label{eqn:cr_newton_update}
        \mat{w} = \mat{w}^* - \left( \mat{H}^r_{\mat{w}^*} + \alpha L \mat{I} \right)^{-1} \mat{g}^r_{\mat{w}^*}\ .
    \end{equation}
    We provide the pseudocode for CureNewton in Appendix~\ref{app:cure_newton}.
    
    \begin{remark}
        CureNewton's method improves the convergence rate to the minimizer of $\cL_{D_r}(\mat{w})$ and thus reduces the number of optimization iterations to erase data.
        In the existing Newton's method, the non-regularized surrogate function (Eq.~\ref{eqn:quadratic_approximation}) only approximates $\cL_{D_r}(\mat{w})$ well in the locality of $\mat{w}^*$, i.e. $\lVert \mat{w} - \mat{w}^* \rVert \leq \epsilon$. However, the model parameters may need to be updated drastically if influential samples, such as an entire class in the training set, are erased. When $\mat{w}^*$ is far from $\mat{w}$, Newton's method may not converge. 
        In contrast, as stated in Lemma 2 of \citet{nestero06:cubic-newton}, our regularized surrogate function (Eq.~\ref{eq:cubic_approx_true}) approximates and upper bounds $\cL_{D_r}(\mat{w})$ well globally, i.e. $\cL_{D_r}(\mat{w}) \leq \widetilde{\cL}_{D_r}(\mat{w}),\ \forall \mat{w} \in \bR^d$. As a result, CureNewton's method provides global convergence to the minimizer of $\cL_{D_r}(\mat{w})$ (with quadratic rate) regardless of $\mat{w}^*$.
    \end{remark}
    
    \begin{remark}
        The optimized damping factor $\alpha$ (scaled by $L$, Eq.~\ref{eqn:cr_newton_update}) benefits the explainability of CureNewton's method for unlearning trained NNs.
        When $\alpha$ has converged (i.e., $v_l'(\alpha) \rightarrow 0$), $\alpha \approx \lVert \Delta_\alpha \rVert = \lVert \mat{w} - \mat{w}^* \rVert$ and can be interpreted as the $\ell_2$-distance between the model parameters before and after learning. 
        This natural interpretation of $\alpha$ enables us to validate unlearning results (w.r.t. our intuition), which will be demonstrated in our experiment (Sec.~\ref{exp:ablation}).
    \end{remark}

    {\bf Practical considerations.}
    Given $d$ model parameters and $n$ data samples, CureNewton incurs a relatively high space complexity $O(d^2)$ for storing the Hessian matrix and time complexity $O(nd^2 + d^3)$ to form and inverse the Hessian matrix.
    We reduce the space complexity to $O(d)$ and time complexity to $O(kd)$ with our stochastic CureNewton's method (SCureNewton) which adapts stochastic cubic regularization from \citet{tripuraneni18:stochastic-cubic}, where $k$ hides multiplicative number of stochastic optimization iterations.
    Particularly, SCureNewton involves optimization of the stochastic regularized surrogate function (Eq.~\ref{eq:cubic_approx_true} measured on random samples of the training dataset) for $k_{\text{outer}}$ iterations. In each iteration, we sample two independent batches to compute the stochastic gradient and stochastic Hessian-vector product (HVP), whose computational cost can be reduced to that of the stochastic gradient~\citep{pearlmutter94:fast-hessian}.
    Subsequently, gradient descent is used to optimize the stochastic problem for $k_{\text{inner}}$ steps, where each step involves recalculation of the stochastic HVP. Hence, $k = k_{\text{inner}} * k_{\text{outer}}$. We provide the pseudocode for SCureNewton in Appendix~\ref{app:scure_newton}.
    
    

    \section{Experiments}
    \label{sec:exp}

\begin{table*}[t]
    \centering
    \resizebox{0.95\textwidth}{!}{
    \begin{tabular}{c|cccccccc}
        \toprule
         \multirow{2}{*}{$\mathcal{M}$} & \multicolumn{4}{c}{\textbf{Selective Unlearning (Random Instance Unlearning)}} & \multicolumn{4}{c}{\textbf{Class Unlearning}} \\
         \cmidrule(l){2-5} \cmidrule(l){6-9}
         & $D_e$ Acc. ($\rightarrow$) & $D_r$ Acc. ($\uparrow$) & $D_{test}$ Acc. ($\uparrow$) & JS Div. ($\downarrow$) & $D_e$ Acc. ($\rightarrow$) & $D_r$ Acc. ($\uparrow$) & $D_{test}$ Acc. ($\uparrow$) & JS Div. ($\downarrow$) \\ \midrule
         Retraining (reference) & 85.43$\pm$0.39	& 87.40$\pm$0.62	& 84.88$\pm$0.54  & 0.0$\pm$0.0 & 0.0$\pm$0.0	& 91.20$\pm$0.97	& 81.29$\pm$0.77 & 0.0$\pm$0.0  \\ \midrule
         Original & 88.85$\pm$0.20	& 88.89$\pm$0.05	& \underline{87.84}$\pm$0.27 & \textbf{0.001}$\pm$0.0 & 85.95$\pm$0.92	& 89.85$\pm$1.01	& \underline{88.38}$\pm$0.67    & 0.021$\pm$0.001  \\
         Rand. Lbls. & 88.31$\pm$0.49	& 88.30 $\pm$0.43	& 87.36$\pm$0.58 & \textbf{0.001}$\pm$0.0 & 9.78$\pm$2.57	& 69.56$\pm$17.53	& 63.03$\pm$15.27   & 0.010$\pm$0.002 \\
         GD   & 89.35$\pm$0.20	& \textbf{89.46}$\pm$0.19	& \textbf{88.34}$\pm$0.23 &   \textbf{0.001}$\pm$0.0    & 84.69$\pm$0.94	& 90.11$\pm$0.92	& \textbf{88.46}$\pm$0.53    &  0.021$\pm$0.001 \\
         GA    & 84.69$\pm$1.26	& 84.65$\pm$1.20		& 83.84$\pm$1.35 &  \underline{0.002}$\pm$0.0   & 7.13$\pm$0.90	& 72.69$\pm$17.09	& 65.54$\pm$15.03  & 0.005$\pm$0.003  \\
         NTK    & 88.81$\pm$0.15	& \underline{89.03}$\pm$0.06	& 87.77$\pm$0.18 &   \textbf{0.001}$\pm$0.0  & 32.38$\pm$20.15   & \underline{90.34}$\pm$1.0     & 83.63$\pm$2.78  & 0.007$\pm$0.003 \\ 
         PINV-Newton & 9.74$\pm$4.11	& 9.83$\pm$4.21	& 9.48$\pm$4.08  & 0.026$\pm$0.003 &  1.43$\pm$2.49	& 8.85$\pm$2.43	& 8.39$\pm$2.08 & 0.032$\pm$0.001  \\	
         Damped-Newton & 8.47$\pm$1.31 & 8.77$\pm$0.94 & 8.88$\pm$1.10 & 0.029$\pm$0.0 & \underline{0.52}$\pm$0.90	& 10.07$\pm$1.39		& 9.28$\pm$0.95 &  0.024$\pm$0.015 \\ 
         SCRUB & 83.95$\pm$1.29	& 84.64$\pm$1.297	& 83.29$\pm$1.31 &  \textbf{0.001}$\pm$0.0 & \textbf{0.0}$\pm$0.0 & \textbf{92.41}$\pm$0.57 & 82.38$\pm$0.38  & \textbf{0.001}$\pm$0.0  \\ \midrule
         CureNewton (ours) & \underline{86.07}$\pm$0.24	& 86.38$\pm$0.57	& \underline{85.19}$\pm$0.09 & \underline{0.002}$\pm$0.0 & 1.37$\pm$0.78	& 88.65$\pm$2.03	& 79.15$\pm$1.95 & \underline{0.002}$\pm$0.001  \\
         SCureNewton (ours) & \textbf{85.93}$\pm$0.54	& 86.26$\pm$0.72	& 85.04$\pm$0.51 & \textbf{0.001}$\pm$0.0 & \underline{0.52}$\pm$0.45	& 89.87$\pm$0.80	& 80.00$\pm$0.66    & \textbf{0.001}$\pm$0.0 \\
         \bottomrule
    \end{tabular}
    }
    \caption{Performance comparison between CureNewton, its stochastic variant SCureNewton, and other tested baselines in batch unlearning settings on CNN $\times$ FashionMNIST (for 3 random runs).
    ``$\rightarrow$'' indicates closer to Retraining is better; ``$\uparrow$'' indicates higher is better; ``$\downarrow$'' indicates lower is better.
    We use boldface to denote best results and underline to denote second best results.
    }
    \label{table:batch_unlearning}
    \vspace{-2mm}
\end{table*}
    
\subsection{Evaluation Metrics} 
We use the following criteria to measure the performance of the unlearning algorithms:

{\bf Erasing quality:} 
We evaluate the erasing quality using two metrics: the accuracy of the model after unlearning on the erased set ($D_e$ Acc.) and the Jensen–Shannon Divergence (JS Div.) between the predictions of the unlearned model and the retrained model on $D_e$ \citep{chundawat23:bad-teaching}. Ideally, the unlearned model should match the retrained model's accuracy on $D_e$ and produce a predictive distribution similar to that of the retrained model on $D_e$.



{\bf Model performance:} 
We compare the accuracy of the model after unlearning on the retained set ($D_r$ Acc.) and a fixed test set ($D_{test}$ Acc.). The former is to measure the learning performance of the unlearned model, while the latter is to measure its generalizability on the hold-out dataset.

{\bf Efficiency:} 
We compare the running time to unlearn a batch of requested data points (batch unlearning). For experiments that involve multiple unlearning rounds (i.e., sequential unlearning), we report the average results per round.
Ideally, unlearning algorithms should be faster than retraining the model from scratch.




\subsection{Baselines for Comparison}

We use the following baselines for comparison:
(1) {\bf Retraining} is an exact unlearning algorithm that retrains the entire model from scratch without the erased set;
(2) {\bf Original} is the model prior to unlearning;
(3) {\bf Random Labels} (Rand. Lbls.) randomly assigns new labels to the erased set (from the retained classes) and finetunes the model on the newly labeled set;
(4) {\bf Gradient Descent} (GD) fine-tunes the model following the descent direction of the gradient (first-order information) on the retained set;
(5) {\bf Gradient Ascent} (GA) maximizes the loss, equivalently follows the ascent direction of the gradient on the erased set;
(6) {\bf Neural Tangent Kernel-based} (NTK)~\citep{golatkar20:forget-ntk} uses NTK matrix to linearize the model outputs and derives a one-shot unlearning update;
(7) {\bf SCRUB}~\citep{kurmanji23:unbounded-unlearn} adopts a teacher-student framework, in which the student model inherits the teacher model's knowledge on the retrained set while diverging from the teacher's behavior on the erased set;
(8) {\bf PINV-Newton} is Newton's method that takes the pseudoinverse of the Hessian matrix to resolve degeneracy (Sec.~\ref{sec:baselines_degeneracy});
(9) {\bf Damped-Newton} is Newton's method that adds a small damping diagonal to the Hessian before inversed (Sec.~\ref{sec:baselines_degeneracy}); we set the default damping factor $\gamma = 10^{-3}$.

\subsection{Batch Unlearning}
\label{sec:exp:batch_unlearning}

Following the work of~\citet{kurmanji23:unbounded-unlearn}, we consider two unlearning scenarios: {\it selective unlearning}, where a random subset of the training data is erased, and {\it class unlearning}, where an entire training class is erased. We refer to both of these as {\it batch unlearning} to suggest that multiple data points are unlearned simultaneously in one batch, whether it involves a random subset or an entire class.\footnote{The term is used to distinguish with our next experiment about sequential unlearning, where data points are unlearned one after another in a specific order.} We argue that batch unlearning is a practical setting because a data owner may request the removal of multiple data points at once, or a specific class may no longer be relevant (e.g. due to domain shift), or a model owner may choose to perform unlearning periodically to minimize computational overhead.

To set up, we use the popular image dataset FashionMNIST~\citep{xiao2017fashionmnist} with $60,000$ training and $10,000$ test images of $10$ fashion categories. 
The model to be unlearned is a convolutional neural network (CNN) comprising two $3 \times 3$ convolutional layers, each followed by a $2 \times 2$ max-pool layer for feature extraction and two fully connected layers ($16$ hidden units) with ReLU activation for classification. 
This experiment is small-scale, unlike the large-scale experiment that follows in Sec.~\ref{sec:exp:sequential_unlearning}, to enable comparisons with NTK and to assess the sensitivity of degeneracy baselines (i.e., PINV-Newton and Damped-Newton), which forms and inverses the full Hessian matrix.
$D_e$ consists of either a random subset of 80\% training samples\footnote{We remove a large subset to observe non-trivial changes in model outputs and performance.} or the entire class $0$ (10\% training samples).

\begin{table*}[t]
    \centering
    \resizebox{0.95\textwidth}{!}{
    \begin{tabular}{c|cccccccc}
        \toprule
         \multirow{2}{*}{$\mathcal{M}$} & \multicolumn{4}{c}{\textbf{Llama-2 $\times$ AG-News}} & \multicolumn{4}{c}{\textbf{ResNet18 $\times$ CIFAR-10}}  \\
         \cmidrule(l){2-5} \cmidrule(l){6-9} 
         & $D_e$ Acc. ($\rightarrow$) & $D_r$ Acc. ($\uparrow$) & $D_{test}$ Acc. ($\uparrow$) & JS Div. ($\downarrow$) & $D_e$ Acc. ($\rightarrow$) & $D_r$ Acc. ($\uparrow$) & $D_{test}$ Acc. ($\uparrow$) & JS Div. ($\downarrow$) \\ \midrule
         Retraining (reference) & 0.0$\pm$0.0   & 95.20$\pm$1.28   & 70.62$\pm$1.04  &  0.0$\pm$0.0  & 0.0$\pm$0.0  & 85.85$\pm$1.45    & 75.16$\pm$2.01   & 0.0$\pm$0.0 \\ \midrule
         Original  & 98.24$\pm$2.17   & 94.59$\pm$0.67   & \textbf{94.60}$\pm$0.10  & 0.014$\pm$0.008 & 96.55$\pm$0.69    & 84.62$\pm$0.12  & 82.77$\pm$0.28   & 0.027$\pm$0.004  \\
         Rand. Lbls. & \textbf{0.0}$\pm$0.0     & 79.19$\pm$9.49   & 59.01$\pm$7.06 & \underline{0.019}$\pm$0.025 & \textbf{0.0}$\pm$0.0  & 18.72$\pm$2.08   & 18.07$\pm$2.24   & \textbf{0.011}$\pm$0.001  \\
         GD     & 59.21$\pm$4.66   & \underline{95.64}$\pm$0.81   & \underline{85.42}$\pm$1.38  & 0.045$\pm$0.017    & 90.13$\pm$11.18    & \textbf{89.65}$\pm$1.54   & \textbf{87.06}$\pm$0.47   & 0.020$\pm$0.008  \\
         GA     & \textbf{0.0}$\pm$0.0    & 33.33$\pm$0.0   & 25.00$\pm$0.0 &  0.038$\pm$0.016    & \underline{5.55}$\pm$9.62      & 25.25$\pm$1.78   & 22.64$\pm$2.36   &  0.028$\pm$0.003  \\
         SCRUB    & \underline{9.68}$\pm$4.07	& \textbf{96.48}$\pm$0.82	& 73.14$\pm$1.33 &  \textbf{0.018}$\pm$0.012  & \textbf{0.0}$\pm$0.0	& \underline{88.42}$\pm$0.08	& \underline{77.52}$\pm$0.22 & \underline{0.019}$\pm$0.012  \\
         \midrule	
         SCureNewton (ours) & 16.45$\pm$15.01	& 89.03$\pm$9.59	& 70.32$\pm$8.92  & 0.023$\pm$0.007 & 17.01$\pm$9.70 & 84.06$\pm$2.33  & 75.27$\pm$2.20  & 0.020$\pm$0.017  \\
         \bottomrule
    \end{tabular}
    }
    \caption{Performance comparison between SCureNewton and other tested baselines in large-scale sequential unlearning setting on Llama-2 $\times$ AG-News and ResNet18 $\times$ CIFAR-10 (for $3$ random runs). Results are reported at the last unlearning round.
    }
    \label{table:sequential_unlearning}
\end{table*}

Table~\ref{table:batch_unlearning} provides our results for batch unlearning on the FashionMNIST dataset.
{\it In the selective unlearning} (random instance unlearning) setting, Retraining achieves a $3.42\%$ accuracy lower than Original on $D_e$. Similarly, we observe a decrease of $2.78\%$ and $2.92\%$ in $D_e$ accuracy for CureNewton and SCureNewton, respectively. Compared to other baselines, CureNewton and SCureNewton maintain the closest and second closest $D_e$ accuracy and JS divergence to Retraining, indicating {\bf effective unlearning} of the erased data points. Additionally, both methods can {\bf preserve model performance} post-unlearning: CureNewton obtains 86.38\% $D_r$ accuracy and $85.19\%$ $D_{test}$ accuracy, while SCureNewton obtains $86.26\%$ $D_r$ accuracy and $85.04\%$ $D_{test}$ accuracy. 
As anticipated, both methods significantly outperform other degeneracy baselines (i.e. PINV-Newton and Damped-Newton), which severely degrade model performance on $D_e$, $D_r$, and $D_{test}$. A closer analysis reveals that the $\ell_2$-norm of the update for PINV-Newton is $3708.78\pm3364.67$, Damped-Newton is $838.68\pm742.96$, while CureNewton incurs $0.36\pm0.07$, and SCureNewton incurs $0.38\pm0.05$. These results affirm the numerical sensitivity of the baselines to the degenerate Hessian mentioned in Sec.~\ref{sec:baselines_degeneracy}. 
{\it In the class unlearning setting}, Retraining achieves $0\%$ $D_e$ accuracy, which is reasonable because the model is no longer trained to classify the erased class. Similar to selective unlearning setting, CureNewton and SCureNewton demonstrate {\bf significant erasing quality} with $1.37\%$ and $0.52\%$ $D_e$ accuracy, respectively, along with a low JS divergence to Retraining {\bf without sacrificing model performance}: CureNewton obtains $88.65\%$ $D_r$ accuracy and $79.15\%$ $D_{test}$ accuracy, while SCureNewton obtains $89.87\%$ $D_r$ accuracy and $80.00\%$ $D_{test}$ accuracy. We also observe disastrous effects on model performance not only in degeneracy baselines but also from GA and Rand. Lbls..


Overall, both CureNewton and SCureNewton demonstrate a strong ability to unlearn random samples and classes in batch unlearning settings, while maintaining a decent model performance post-unlearning. As such, both methods are strong competitors to the state-of-the-art method SCRUB, with the added advantage of being theoretically supported. 
We also show that SCureNewton offers an economical alternative to CureNewton's method, which allows us to evaluate on a larger scale in the next experiment.

\paragraph{Discussion on Membership Inference Attack (MIA).} Unlearning can also be evaluated from the privacy perspective. In Appendix~\ref{app:mia}, we compare the accuracy of the standard MIA, which performs binary classification on the losses of the unlearned model on $D_e$ and $D_{test}$, following the work of~\citet{kurmanji23:unbounded-unlearn}. Ideally, an attacker cannot distinguish samples from $D_e$ and $D_{test}$ after unlearning, where $D_e$ and $D_{test}$ samples are drawn from the same distribution, e.g., samples of the same class.
Hence, the ideal MIA accuracy is around $50\%$.

\subsection{Sequential Unlearning}
    \label{sec:exp:sequential_unlearning}
    
    \begin{figure}[t!]
        \centering
        \includegraphics[width=\linewidth]{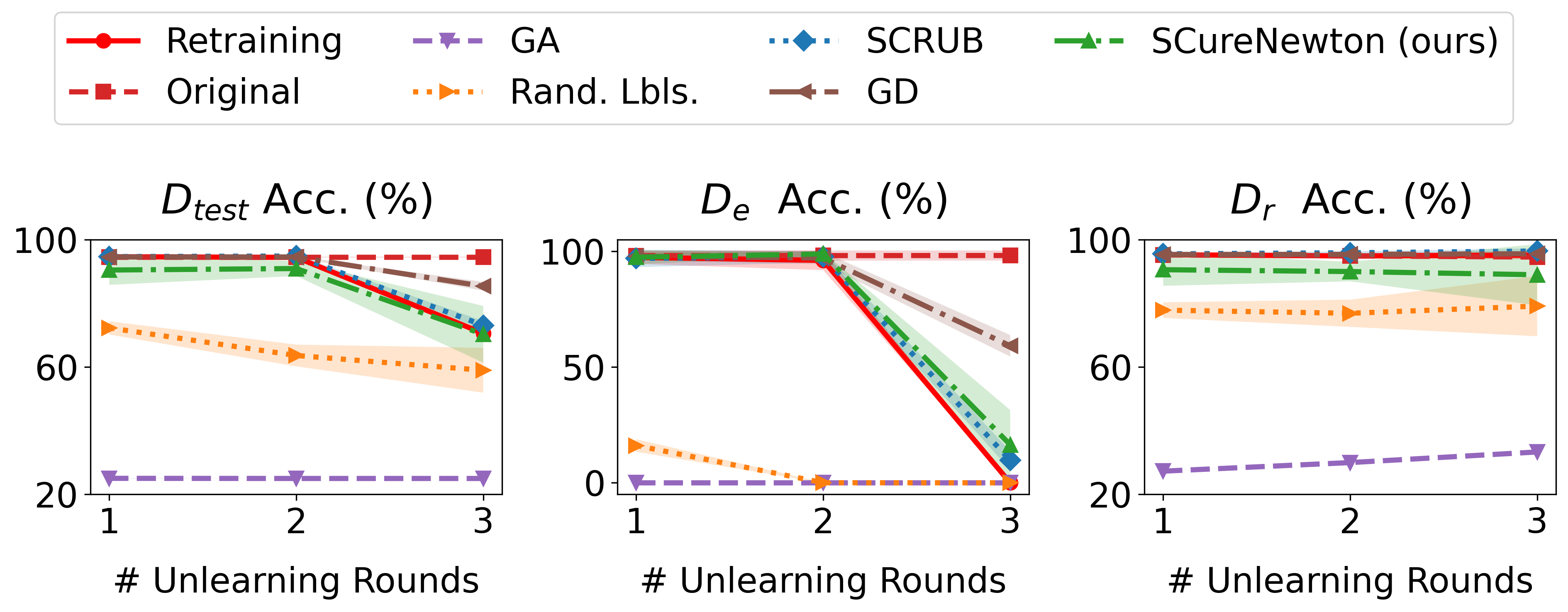} \\
        \includegraphics[width=\linewidth]{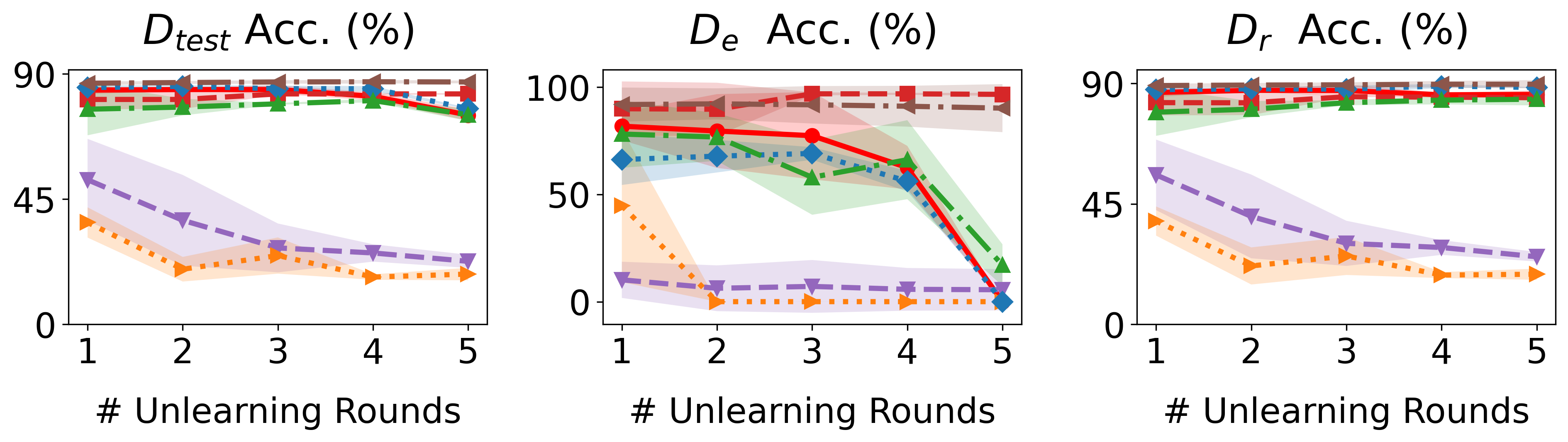}
        \caption{Performance comparison in the sequential unlearning settings on different datasets and models. Top row: Llama-2 $\times$ AG-News. Bottom row: ResNet18 $\times$ CIFAR-10. }
        \label{fig:sequential_unlearning}
        \vspace{-2mm}
    \end{figure}
    
    When unlearning is not handled properly for a single request, errors can accumulate and deteriorate model performance over time. To demonstrate this, we conduct worst-case experiments involving unlearning a model multiple times, or {\it sequential unlearning}. Specifically, we perform \textit{class unlearning} over $n$ unlearning rounds (hence, similar to Sec.~\ref{sec:exp:batch_unlearning}, the final model is expected to perform badly on the erased class); in each unlearning round, we perform batch unlearning on a random subset of the entire class, such that each round (except the last round) can be viewed as a \textit{selective unlearning} scenario as well. This design is appealing because it unveils the long-term effectiveness of the unlearning algorithms and helps us identify catastrophic forgetting phenomena more easily. Besides, we argue that sequential unlearning appropriately reflects real-world scenarios when unlearning requests may come in streams and/or when resource constraints (e.g. data accessibility) limit the amount of data that can be unlearned at a certain time. 
    

    In contrast to small-scale models in Sec.~\ref{sec:exp:batch_unlearning}, our experiments in this section are carried on two popular large-scale models for images and text: Llama-2-7B~\citep{touvron2023llama2} and ResNet18~\citep{he16:resnet}. Specifically, we incorporate Low-Rank Adaption (LoRA)~\citep{hu2021lora} to efficiently fine-tune a pre-trained Llama-2-7B~\footnote{\url{https://huggingface.co/meta-llama/Llama-2-7b-hf}.} on the AG-News dataset, which consists of $120,000$ training and $7,600$ test samples in $4$ balanced classes. We fine-tune a ResNet18 (pre-trained on ImageNet-1K dataset) on the CIFAR-10 dataset~\citep{krizhevsky2009learning} comprising of $50,000$ training and $1000$ test images in $10$ balanced classes. The number of unlearning rounds is set to $3$ for the AG-News experiment and $5$ for the CIFAR-10 experiment. Note that for these models, many baseline unlearning algorithms (i.e., NTK, PINV-Newton, and Damped-Newton) and our CureNewton are not applicable as they poorly scale to large model sizes.

    Fig.~\ref{fig:sequential_unlearning} and Table~\ref{table:sequential_unlearning} show our results in the sequential unlearning settings. We observe that our SCureNewton achieves $D_e$ accuracy close to Retraining on both datasets while maintaining decent accuracy on $D_{test}$ and $D_r$. 
    On an absolute scale, the final models using Retraining show $0\%$ $D_e$ accuracy, whereas those using SCureNewton achieve around $16-17\%$ $D_e$ accuracy. 
    This underscores SCureNewton's ability to {\bf achieve high erasing quality while preserving the performance of the post-unlearning model, even after multiple unlearning requests}. 
    When compared to other baseline methods, SCureNewton falls short of the state-of-the-art method SCRUB by $6\%$ and $17\%$ on $D_e$ accuracy. Despite this, our SCureNewton offers advantages in theoretical support and efficiency, as we will show in Sec.~\ref{sec:exp:efficiency}.
    When compared to GD, SCureNewton exhibits significantly better erasing quality due to leveraging second-order information (the Hessian) in addition to first-order information (the gradient), which is particularly beneficial for unlearning large-scale models. 
    On the other hand, catastrophic forgetting can be pronouncedly observed in other baseline methods (i.e., GA and Rand. Lbls.), indicating their inadequacy in performing unlearning in long-term settings.

    \begin{table}[t]
        \centering
        \resizebox{0.98\linewidth}{!}{
        \begin{tabular}{c|cccc}
            \toprule
              Dataset & \textbf{FashionMNIST} & \textbf{AG-News} & \textbf{CIFAR-10}  \\ \midrule
              Model & 2-layer CNN & Llama-2-7B (+LoRA) & ResNet18  \\ \midrule
             \# Model Parameters & 20,728 & 1,064,960 & 11,173,962  \\ 
             \midrule
             Retraining & 61.20$\pm$8.70 & 4792.44$\pm$145.90 & 124.51$\pm$10.95  \\ 
             SCRUB & \textbf{23.33}$\pm$0.43 & \underline{6796.16}$\pm$160.11 & \underline{72.39}$\pm$4.93  \\
             CureNewton (ours) & 6355.31$\pm$127.31  & NA & NA \\ 
             SCureNewton (ours) & \underline{35.54}$\pm$6.73 & \textbf{85.26}$\pm$18.23 & \textbf{41.79}$\pm$0.94  \\
             \bottomrule
        \end{tabular}
        }
        \caption{Running time comparison (in seconds) of the best performing unlearning algorithms across different datasets and models (from $3$ random runs). 
        }
        \label{tab:efficiency}
        \vspace{-3mm}
    \end{table}
    
    %
    %
    
    \subsection{Efficiency}
    \label{sec:exp:efficiency}
    In Sections~\ref{sec:exp:batch_unlearning} and \ref{sec:exp:sequential_unlearning}, we have shown that CureNewton, SCureNewton, and SCRUB can achieve impressive unlearning ability across various datasets and models.
    To understand the efficiency of these methods, Table~\ref{tab:efficiency} shows the average running time comparison to unlearn one batch of erased data points against Retraining.
    As anticipated, despite its strong unlearning performance on FashionMNIST, CureNewton is the least efficient algorithm due to the extensive computation required to calculate and invert the full degenerate Hessian matrix.
    SCureNewton provides an economical alternative to CureNewton and can perform unlearning much faster than Retraining, especially on large models such as Llama-2.
    It's worth mentioning that we only train/retrain Llama-2 on AG-News for $1$ epoch, hence SCRUB incurs a higher computational cost than Retraining because SCRUB runs $1$ epoch to maximize the divergence on $D_e$ and another $1$ epoch to minimize the predictive loss on the $D_r$. In contrast, SCureNewton \textbf{samples only a few batches} from $D_r$, yet still achieves non-trivial unlearning performance compared to SCRUB.
    Therefore, we argue that SCureNewton is an effective and efficient unlearning algorithm that can leverage second-order information in a principled way to unlearn modern NNs.

    \subsection{Ablation: $\alpha$ Dynamics}
        
    \label{exp:ablation}

    \begin{figure}
        \centering
        \includegraphics[width=0.5\linewidth]{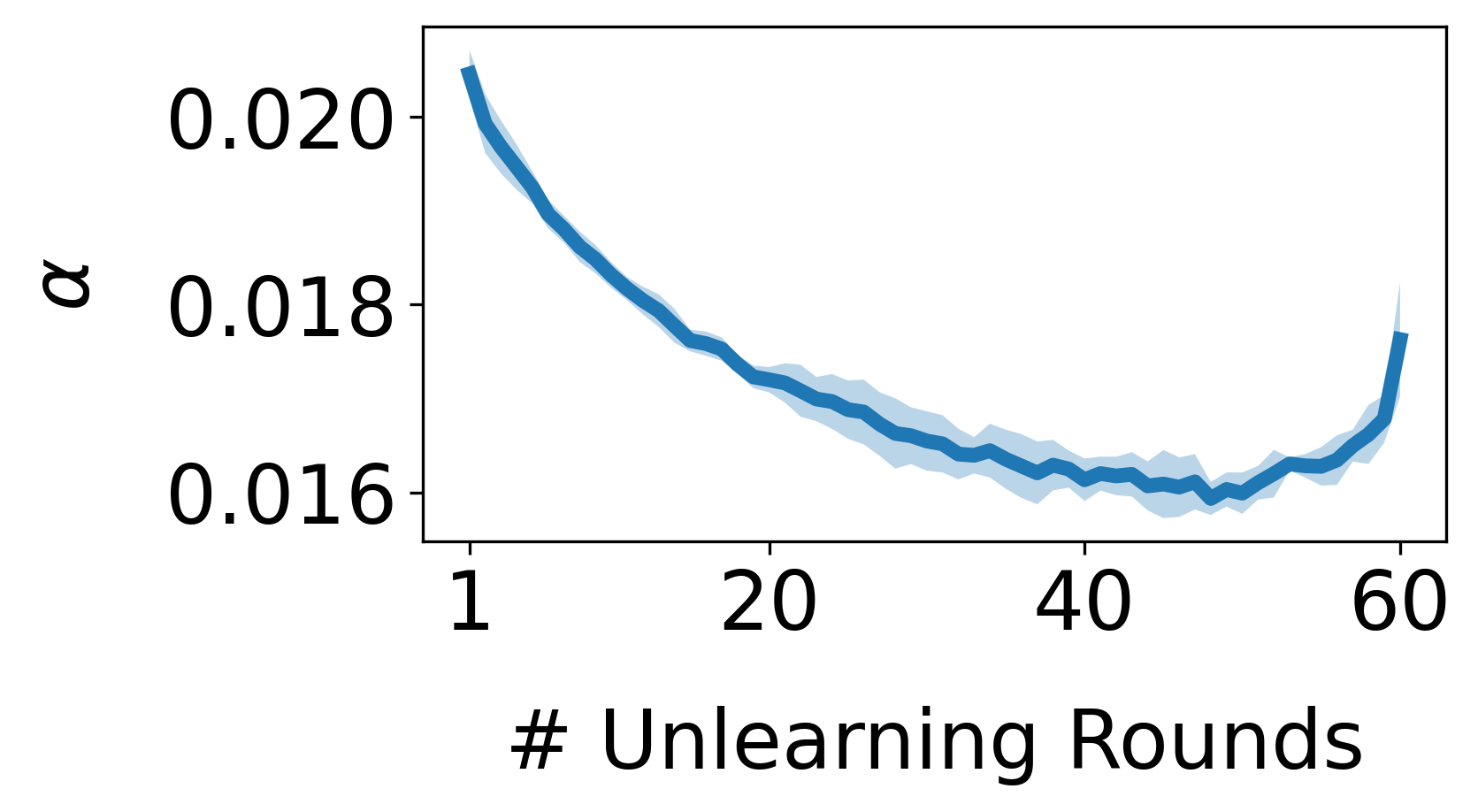}
        \caption{$\alpha$ dynamics in CureNewton's method during sequential unlearning on FashionMNIST (for $3$ random runs).}
        \label{fig:alpha}
        \vspace{-3.4mm}
    \end{figure}
    In CureNewton, we have introduced a dual variable $\alpha$ that converges to $\lVert \Delta_\alpha \rVert = \lVert \mat{w} - \mat{w}^* \rVert$, which can be interpreted as $\ell_2$-distance between the model parameters before and after unlearning (Sec.~\ref{sec:method}). In this ablation, we show how $\alpha$ is optimized for different unlearning requests in \textit{sequential unlearning} on FashionMNIST in Fig.~\ref{fig:alpha}.
    On an absolute scale, we observe that $\alpha$ obtains larger values than the default damping factor $\gamma = 10^{-3}$ in Damped-Newton, which effectively prevents the model from making offensively large norm updates.
    During sequential unlearning, $\alpha$ consistently decreases, suggesting that the Hessians require less regularization to become p.d. and well-defined for unlearning.
    The sudden increase of $\alpha$ at the last unlearning round coincides with the unlearning of an entire class, which signifies that the algorithm converges to the parameters further away from the previous parameters. This behavior aligns with our intuition -- there is a tremendous shift in decision boundary, hence a larger norm update, when we remove an entire class. 

    \section{Conclusion}
    \label{sec:conclusion}
    Despite its proven successes on linear models, adapting Newton's method to unlearn trained NNs would encounter non-trivial problems with degenerate Hessians.
    To address these challenges, we have proposed CureNewton's method and its stochastic variant SCureNewton's method to effectively and efficiently unlearn trained NNs.
    The empirical evaluation affirms that our proposed methods can facilitate practical unlearning settings, such as batch unlearning and sequential unlearning for both 
    selective unlearning and class unlearning
    while being theoretically justified and efficient in running time.
    For future work, it would be beneficial to analyze the theoretical guarantee of CureNewton's method for approximate unlearning in the $(\epsilon, \delta)$-certified removal framework~\cite{guo20:certified-removal}
    and continue improving the efficiency of second-order methods for unlearning.

\bibliography{aaai25}

\appendix

\onecolumn

\section{Derivation of Remark 4.1 and 4.2}

    {\bf Observation 1.}
    {\it At convergence, $\mat{H}_{\mat{w}^*}^r \in \bR^{d \times d}$ is degenerate whose eigenvalues (in non-increasing order) are $\{\lambda_1, \ldots, \lambda_k, 0, \ldots, 0\}$ ($k < d$) and the corresponding eigenvectors are $\{\mat{v}_1, ..., \mat{v}_d\}$.}
    \vspace{2mm}

Let $\mat{\Lambda} := \text{diag}(\lambda_1, \ldots, \lambda_k, 0, \ldots, 0) \in \bR^{d \times d}$ be a diagonal matrix of the eigenvalues of $\mat{H}_{\mat{w}^*}^r$.
Let $\mat{Q} := [\mat{v}_1, ..., \mat{v}_d] \in \bR^{d \times d}$ whose columns are the orthornormal eigenvectors of $\mat{H}_{\mat{w}^*}^r$.
We assume that $\mathcal{L}_{D_r}$ is continuous and $\mat{H}_{\mat{w}^*}^r$ is a symmetric matrix.

\subsection{Derivation of Remark 4.1}
\label{sec:proof_rmk_pinv}

{\it Remark} 4.1. Under Obs.~\ref{obs:hessian_eigenspectrum}, $\lVert \Delta \rVert = \sqrt{\sum_{i=1}^k \frac{1}{\lambda_i^2} | \mat{v}_i^\top \mat{g}_{\mat{w}^*}^r |^2}$ where $\Delta = (\mat{H}_{\mat{w}^*}^r)^\dagger \mat{g}_{\mat{w}^*}^r$.

Let us define $\mat{\Lambda}^\dagger := \text{diag}\left( \frac{1}{\lambda_1}, ..., \frac{1}{\lambda_k}, 0, ..., 0 \right)$ to be the pseudoinverse of $\mat{\Lambda}$.
\begin{align}
    (\mat{H}_{\mat{w}^*}^r)^\dagger \mat{g}_{\mat{w}^*}^r
    &= (\mat{Q} \mat{\Lambda} \mat{Q}^{-1})^\dagger \mat{g}_{\mat{w}^*}^r \label{rmk_pinv_s1} \\
    &= (\mat{Q} \mat{\Lambda} \mat{Q}^\top)^\dagger \mat{g}_{\mat{w}^*}^r \label{rmk_pinv_s2} \\
    &= (\mat{Q} \mat{\Lambda}^\dagger \mat{Q}^\top) \mat{g}_{\mat{w}^*}^r \label{rmk_pinv_s3}  \\
    &= \left(\sum_{i = 1}^k \frac{1}{\lambda_i} \mat{v}_i \mat{v}_i ^\top \right) \mat{g}_{\mat{w}^*}^r\ .  \label{rmk_pinv_s4}
\end{align}
Eq.~\ref{rmk_pinv_s1} is the direct application of eigendecomposition to the symmetric matrix $\mat{H}_{\mat{w}^*}^r$.
In Eq.~\ref{rmk_pinv_s2} and \ref{rmk_pinv_s3}, we use the fact that $\mat{Q}^{-1} = \mat{Q}^\top$ for the orthonormal matrix $\mat{Q}$ and $\mat{Q}^\dagger = \mat{Q}^{-1}$ since $\mat{Q}$ is invertible.
Eq.~\ref{rmk_pinv_s4} is equivalent to projecting $\mat{g}_{\mat{w}^*}^r$ onto the subspace spanned by $\{ \mat{v}_1, ..., \mat{v}_k \}$ where $\mat{v}_1 \perp ... \perp \mat{v}_k$ and $\lVert \mat{v}_i \rVert = 1$. Hence, by Pythagorean Theorem, we have $\lVert  (\mat{H}_{\mat{w}^*}^r)^\dagger \mat{g}_{\mat{w}^*}^r \rVert = \sqrt{\sum_{i=1}^k \frac{1}{\lambda_i^2} | \mat{v}_i^\top \mat{g}_{\mat{w}^*}^r |^2}$.

\subsection{Derivation of Remark 4.2}
\label{sec:proof_rmk_damping}

{\it Remark} 4.2. Under Obs.~\ref{obs:hessian_eigenspectrum}, $\lVert \Delta \rVert = \sqrt{\sum_{i=1}^k \frac{1}{(\lambda_i + \gamma)^2} | \mat{v}_i^\top \mat{g}_{\mat{w}^*}^r |^2 + \sum_{j=k+1}^d \frac{1}{\gamma^2} | \mat{v}_j^\top \mat{g}_{\mat{w}^*}^r |^2}$ where $\Delta = (\mat{H}_{\mat{w}^*}^r + \gamma \mat{I})^{-1} \mat{g}_{\mat{w}^*}^r$

Let us define $\mat{\Lambda'} := \text{diag} \left( \lambda_1 + \gamma, ..., \lambda_k + \gamma, \gamma, ..., \gamma \right)$ to be the diagonal matrix of the eigenvalues of $\mat{H}_{\mat{w}^*}^r + \gamma \mat{I}$, which is invertible.
\begin{align}
    (\mat{H}_{\mat{w}^*}^r + \gamma \mat{I})^{-1} \mat{g}_{\mat{w}^*}^r 
    &= (\mat{Q} \mat{\Lambda'} \mat{Q}^{-1})^{-1} \mat{g}_{\mat{w}^*}^r \label{rmk_damping_s1} \\
    &= (\mat{Q} \mat{\Lambda'} \mat{Q}^\top)^{-1} \mat{g}_{\mat{w}^*}^r \label{rmk_damping_s2}  \\
    &= (\mat{Q} [\mat{\Lambda'}]^{-1} \mat{Q}^\top) \mat{g}_{\mat{w}^*}^r \label{rmk_damping_s3} \\
    &= \left( \sum_{i = 1}^k \frac{1}{\lambda_i + \gamma} \mat{v}_i \mat{v}_i^\top + \sum_{j = k+1}^d \frac{1}{\gamma} \mat{v}_j \mat{v}_j^\top \right) \mat{g}_{\mat{w}^*}^r\ . \label{rmk_damping_s4}
\end{align}
Eq.~\ref{rmk_damping_s1} is the direct application of eigendecomposition to the symmetric matrix $\mat{H}_{\mat{w}^*}^r + \gamma \mat{I}$.
In Eq.~\ref{rmk_damping_s2} and \ref{rmk_damping_s3}, we use the fact that $\mat{Q}^{-1} = \mat{Q}^\top$ for the orthonormal matrix $\mat{Q}$.
Similar to Sec.~\ref{sec:proof_rmk_pinv}, we have $\lVert  (\mat{H}_{\mat{w}^*}^r + \gamma \mat{I})^{-1} \mat{g}_{\mat{w}^*}^r \rVert = \sqrt{\sum_{i=1}^k \frac{1}{(\lambda_i + \gamma)^2} | \mat{v}_i^\top \mat{g}_{\mat{w}^*}^r |^2 + \sum_{j=k+1}^d \frac{1}{\gamma^2} | \mat{v}_j^\top \mat{g}_{\mat{w}^*}^r |^2}$.


\section{CureNewton's Method}
\label{app:cure_newton}

\subsection{Relation to Trust-Region Subproblem Solver}
\label{app:trust_region}

In Sec. 5.2, we aim to find $\alpha$ for the dual problem:
\begin{equation*}
\begin{aligned}
    & \sup_{\alpha} & & v_l(\alpha) := -\frac{1}{2} \left\langle \left( \mat{H}_{\mat{w}^*}^r + \alpha L \mat{I} \right)^{-1} \mat{g}_{\mat{w}^*}^r, \mat{g}_{\mat{w}^*}^r \right\rangle - \frac{L}{6} \alpha^3
    \quad \text{s.t.} & & \alpha \in Q = \{\alpha \in \bR: \mat{H}_{\mat{w}^*}^r + \alpha L \mathbf{I} \succ 0, \alpha \geq 0 \}\ .
\end{aligned}
\end{equation*}
To do so, we solve the following first-order optimality condition (note that $\alpha \geq 0$):
\begin{equation}
    v_l'(\alpha) = \frac{L}{2} (\lVert \Delta_\alpha \rVert^2 - \alpha^2) = 0 \\
    \quad \Leftrightarrow\ \lVert \Delta_\alpha \rVert = \alpha\ .
\end{equation}

Let $\lambda_d$ be the smallest (negative) eigenvalue of $\mat{H}_{\mat{w}^*}^r$ and $\epsilon$ be an infinitesimal value such that $\mat{H}_{\mat{w}^*}^r + \alpha_0 L \mat{I}$ is positive definite.
We obtain the smallest $\alpha$ (to achieve a positive definite matrix) by initializing $\alpha_0 = \frac{\max(0, -\lambda_d) + \epsilon}{L}$.
As Remark 4.2 shows that the smaller $\alpha$ (or $\gamma$) gives the larger $\lVert \Delta_\alpha \rVert$, i.e., $\lVert \Delta_\alpha \rVert$ is a strictly decreasing function of $\alpha$
and $\alpha_0$ is the smallest, $\lVert \Delta_0 \rVert$ is the largest norm we can get.
We then consider two cases:
\begin{itemize}
    \item {\bf Case 1:}
    If $\lVert \Delta_{\alpha_0} \rVert > \alpha_0$, we can find $\alpha > \alpha_0$ such that $\lVert \Delta_\alpha \rVert = \alpha$. Often, this is the case for non-convex functions such as NNs. Consequently, Newton's method can be employed to find the root of this problem. However, it is worth noting that $\lVert \Delta_\alpha \rVert = \alpha$ may be ill-defined at some points (see Sec. 7.3 in \citet{jorge06:numerical-optimization}). In such cases, we may prefer solving $\alpha$ in the inverse/secular equation $\frac{1}{\lVert \Delta_\alpha \rVert} = \frac{1}{\alpha}$.
    \item {\bf Case 2:}
    If $\lVert \Delta_{\alpha_0} \rVert < \alpha_0$, the solution is made up by the linear combination of the smallest eigenvector $\mat{v}_d$ (Sec 6.1 in \citet{cartis11:adaptive-cubic}). Hence, we can find $\alpha$ such that $\lVert \Delta_\alpha + \alpha L \cdot \mat{v}_d \rVert = \alpha$.
\end{itemize}

In fact, the procedure we presented above is akin to solving the trust-region subproblem~\cite{jorge06:numerical-optimization}:
\begin{equation*}
    \label{eqn:trust-region-subproblem}
    \min_{\Delta_\alpha \in \bR^{d}} \frac{1}{2} \langle \mat{H}_{\mat{w}^*}^r \Delta_\alpha, \Delta_\alpha \rangle + \langle \mat{g}_{\mat{w}^*}^r, \Delta_\alpha \rangle \quad \text{s.t.} \quad \lVert \Delta_\alpha \rVert \leq \alpha\ ,
\end{equation*}
which can be intuitively understood as optimizing the non-regularized surrogate function (Eq. 1) within a $\ell_2$-norm trust region $\lVert \Delta_\alpha \rVert \leq \alpha$ for some radius $\alpha$.

\subsection{Pseudo-code}
\label{app:pseudo_code}

\begin{algorithm}[H]
    \caption{CureNewton's Update for Unlearning}
    \label{algo:cr_newton_update}
        \begin{algorithmic}[1]
        
        \Statex {\bf Input:} original model parameters: $\mat{w}^*$, retained set: $D_r$, objective function: $\cL$, Hessian Lipschitz constant: $L$, tolerance $\epsilon$, maximal Newton's iterations $T$
        \State Calculate $\mat{g}_{\mat{w}^*}^r = \nabla \cL_{D_r}(\mat{w}^*)$
        \State Calculate $\mat{H}_{\mat{w}^*}^r = \nabla^2 \cL_{D_r}(\mat{w}^*)$
        \State $\Delta_\alpha$, $\alpha =$ \Call{TrustRegionSolver}{$\mat{H}_{\mat{w}^*}^r$, $\mat{g}_{\mat{w}^*}^r$, $L$, $\epsilon$, $T$}
        \State Set $\mat{w} = \mat{w}^* - \Delta_\alpha$
        \Statex {\bfseries Output:} unlearned model parameters: $\mat{w}$, (scaled) damping factor: $\alpha$\ ;
        \Statex
        
        \Function{TrustRegionSolver}{$\mat{H}$, $\mat{g}$, $L$, $\epsilon$, $T$} 
        \Comment{See~\ref{app:trust_region}}
        \State Calculate the smallest eigenvalue $\lambda_d$ and eigenvector $\mat{v}_d$ of $\mat{H}$
        \State Initialize $\gamma_0 = \max(0, -\lambda_d) + \epsilon$

        \State Get triangular matrix $\mat{L} \in \bR^{d \times d}$ from the Cholesky decomposition $\mat{L} \mat{L}^\top = \mat{H} + \gamma_0 \mat{I}$
        \State Set $\alpha = \frac{\gamma_0}{L}$
        \State Get $\Delta_\alpha \in \bR^d$ such that $(\mat{L} \mat{L}^\top)^{-1} \Delta_\alpha = \mat{g}$
        
        \If{$\lVert \Delta_\alpha \rVert > \alpha$} 
            \Comment{Find boundary solution}
            \For{$t = 1 .. T$}
                \State Get $\mat{u} \in \bR^d$ such that $\mat{L} \mat{u} = \Delta_\alpha$
                \State Calculate $v_l'(\alpha) = \frac{1}{\lVert \Delta_\alpha \rVert} - \frac{1}{\alpha}$
                \State Calculate $v_l''(\alpha) = \frac{\lVert \mat{u} \rVert^2}{\lVert \Delta_\alpha \rVert^3} + \frac{1}{\gamma_{t-1} \alpha}$
                \State Set $\gamma_t = \gamma_{t-1} - \frac{v_l'(\alpha)}{v_l''(\alpha)}$
                \State Set $\alpha = \frac{\gamma_t}{L}$
                
                \State Get triangular matrix $\mat{L} \in \bR^{d \times d}$ from the Cholesky decomposition $\mat{L} \mat{L}^\top = \mat{H} + \gamma_t \mat{I}$
                \State Get $\Delta_\alpha \in \bR^d$ such that $(\mat{L} \mat{L}^T)^{-1} \Delta_\alpha = \mat{g}$

                \If{$|\lVert \Delta_\alpha \rVert - \alpha| \leq \epsilon$ }
                    \State {\bf break}\ ;
                \EndIf
            \EndFor
            \State {\bf return} $\Delta_\alpha$, $\alpha$\ ;
        \Else
            \Comment{Find interior solution}
            \If{$|\lVert \Delta_\alpha \rVert - \alpha| \leq \epsilon$ }
                \State {\bf return} $\Delta_\alpha$, $\alpha$\ ;
            \Else
                \State Get $\gamma \in \bR$ such that $\lVert \Delta_\alpha + \gamma \cdot \mat{v}_d \rVert - \alpha = 0$
                \State Set $\Delta_\alpha = \Delta_\alpha + \gamma \cdot \mat{v}_d$
                \State Set $\alpha = \frac{\gamma}{L}$
                \State {\bf return} $\Delta_\alpha$, $\alpha$\ ;
            \EndIf
        \EndIf

        \EndFunction
    \end{algorithmic}
\end{algorithm}

\section{SCureNewton's Method}
\label{app:scure_newton}
\begin{algorithm}[H]
    \caption{SCureNewton's Update for Unlearning}
    \label{algo:scr_newton_update}
        \begin{algorithmic}[1]
        \Statex {\bf Input:} original model parameters: $\mat{w}^*$, retained set: $D_r$, objective function: $\cL$, stochastic Hessian Lipschitz constant: $M$, gradient perturbation $\sigma$, step size $\eta$, outer iterations $k_{\text{outer}}$, inner iterations $k_{\text{inner}}$
        \State Set $\mat{w}_0 = \mat{w}^*$

        \For{$t=1..k_{\text{outer}}$}
            \State Sample $B_1$ and $B_2$ independently from $D_r$
            \State Calculate $\widetilde{\mat{g}}_{\mat{w}_{t-1}}^r = \nabla \cL_{B_1}(\mat{w}_{t-1})$
            \State Calculate $\widetilde{\mat{H}}_{\mat{w}_{t-1}}^r = \nabla^2 \cL_{B_1}(\mat{w}_{t-1})$
            \State $\Delta =$ \Call{DescentCubicSolver}{$\widetilde{\mat{H}}_{\mat{w}_{t-1}}^r$, $\widetilde{\mat{g}}_{\mat{w}_{t-1}}^r$, $M$, $\sigma$, $\eta$, $k_{\text{inner}}$}
            \State Set $\mat{w}_t = \mat{w}_{t-1} - \Delta_a$
        \EndFor
        \State Set $\mat{w} = \mat{w}_T$
        \Statex {\bf Output:} unlearned model parameters: $\mat{w}$\ ;
        
        \State

        \Function{DescentCubicSolver}{$\mat{H}$, $\mat{g}$, $M$, $\sigma$, $\eta$, $k_{\text{inner}}$}
            \State Set $R_c = -\frac{\mat{g}^T \mat{H} \mat{g}}{M \lVert \mat{g} \rVert^2} + \sqrt{\left(\frac{\mat{g}^T \mat{H} \mat{g}}{M \lVert \mat{g} \rVert^2}\right)^2 + \frac{2 \lVert \mat{g} \rVert}{M}}$
            \State Set $\Delta_0 = -R_c \frac{\mat{g}}{\lVert \mat{g} \rVert}$
            \State Perturb $\mat{g}$ to obtain $\mat{g}' = \mat{g} + \sigma \boldsymbol{\xi}$ where $\boldsymbol{\xi}$ is sampled on the $d$-dimensional unit sphere
            \For{$i = 1..k_{\text{inner}}$}
                \State $\Delta_{i} = \Delta_{i-1} - \eta \left( \mat{H} \cdot \Delta_{i-1} + \mat{g}' + M \lVert \Delta_{i-1} \rVert \Delta_{i-1} \right)$ 
            \EndFor
            \State {\bf return} $\Delta_{k_{\text{inner}}}$\ ;
        \EndFunction
        \end{algorithmic}
\end{algorithm}

\section{Detailed Experimental Setup}
\label{app:exp_detail}

We conduct our experiments on NVIDIA L40 and H100 GPUs. 
Evaluation is averaged across 3 random seeds $\{ 5, 1, 2\}$. 
The setup of our experiments (by datasets) is detailed below.

\textbf{FashionMNIST.}
The dataset contains $28 \times 28$ grayscale article images belonging to $10$ classes. The training set contains $60,000$ samples, and the test set contains $10,000$ samples.
We use a 2-layer CNN with a convolutional layer (kernel size $3$ and maps to $8$ channels) and a fully connected layer ($10$ hidden units). The model is trained using Adam optimizer with batch size $64$, $15$ training epochs, learning rate $0.01$ decayed at rate $0.5$ every $5000$ step, and weight decay $0.005$.
To unlearn with CureNewton, we set $L = 5$. 
To unlearn with SCureNewton, we use $M = 1$, gradient sample size $128$, Hessian sample size $64$, $20$ outer iterations, $5$ inner iterations with step size $0.01$.

\textbf{CIFAR-10.}
The dataset contains $32 \times 32$ color images in $10$ classes. The training set contains $50,000$ samples, and the test set contains $10,000$ samples.
We train a ResNet18~\citep{he16:resnet} using Adam optimizer with batch size $100$, $10$ training epochs, learning rate $0.001$ decayed at rate $0.5$ every $5000$ step, and weight decay $10^{-4}$. 
To unlearn with SCureNewton, we set $M = 50$, gradient sample size $40$, Hessian sample size $20$, with $10$ outer iterations and $3$ inner iterations.
In class-level experiments, we pick all samples belonging to class $5$ to remove.

\textbf{AG-News.}
The dataset contains news titles and descriptions in $4$ topics. The training set contains $30,000$ samples and the test set contains $1,900$ samples.
We finetune the bfloat16-pretrained Llama-2-7B model from Hugging Face\footnote{\url{https://huggingface.co/meta-llama/Llama-2-7b-hf}.} using LoRA ($r = 2$, $\alpha = 2$) with learning rate $2\times 10^{-4}$ and weight decay $0.01$.
To unlearn with SCureNewton, we set $M = 50$, gradient sample size $20$, Hessian sample size $10$, $10$ outer iterations, and $3$ inner iterations.

\section{Evaluation with Membership Inference Attack (MIA)}
\label{app:mia}

As mentioned in Sec. 6.3, we compare the accuracy of the standard MIA following the work of~\citet{kurmanji23:unbounded-unlearn}, which performs binary classification based on the losses of the unlearned model on $D_e$ and $D_{test}$. Table~\ref{table:sequential_unlearning_MIA} (expansion of the CIFAR-10 results in Table 2 in the main paper) shows the performance and MIA accuracy on CIFAR-10 experiments. However, we observe that the standard MIA accuracy based on losses of the unlearned models is not informative, which we hypothesize is due to the following factors:

\textbf{Overfitting.} As mentioned in~\citet{kurmanji23:unbounded-unlearn}, the original model in their work has overfitted more than a state-of-the-art CIFAR model would, and their MIA is performed on the overfitted original and unlearned models. However, the original model in our setting is more generalized (from Table~\ref{table:sequential_unlearning_MIA}, $D_{test}$ and $D_r$ accuracy are close to each other with a difference of $1.85\%$), and the MIA accuracy on our non-overfitted models is less informative. While studies have shown that there are connections between overfitting and privacy leakage~\citep{shokri2017membership, yeom2018privacy}, in our setting, when the original model generalizes well on $D_{test}$, the losses on $D_e$ and $D_{test}$ would be less distinguishable, which leads to close-to-$50\%$ MIA accuracy for both original and unlearned models. Therefore, we conduct a new experiment by training the original CIFAR-10 model for $50$ epochs to obtain an overfitted model (from Table~\ref{table:mia_overfit}, $D_{test}$ accuracy is much lower than $D_r$ accuracy with a difference of $7.17\%$) and perform unlearning on the overfitted model. However, in practice, an overfitted model is less preferable than those with better generalization abilities. Table~\ref{table:mia_overfit} shows the performance and MIA results for different unlearning algorithms on the overfitted CIFAR-10 model, where the MIA accuracy is more informative. Based on the results, our SCureNewton successfully decreases MIA accuracy by $4.34\%$. While there remains a gap between SCureNewton and the state-of-the-art method SCRUB, our SCureNewton offers advantages in theoretical support and efficiency.

\begin{figure}[t!]
    \centering
    \includegraphics[width=0.3\linewidth]{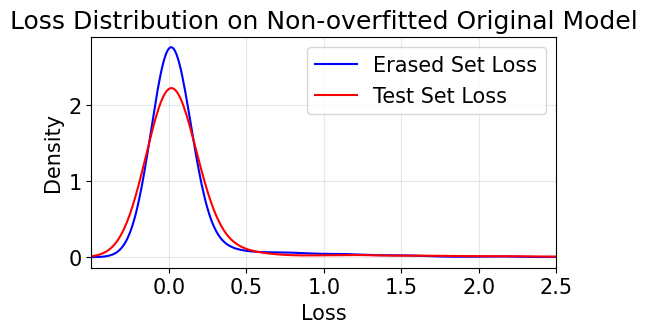} 
    \includegraphics[width=0.32\linewidth]{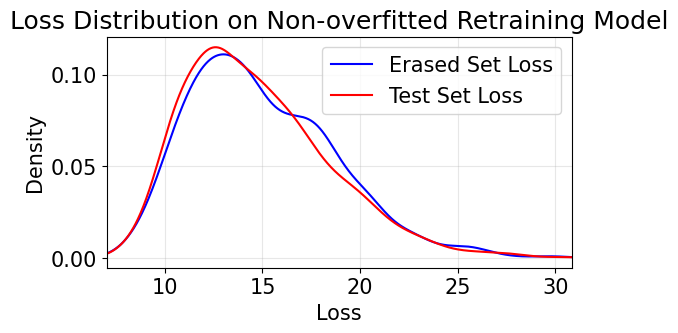} \\
    \includegraphics[width=0.3\linewidth]{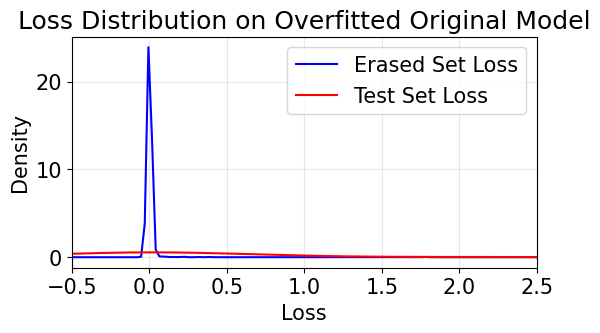} 
    \includegraphics[width=0.32\linewidth]{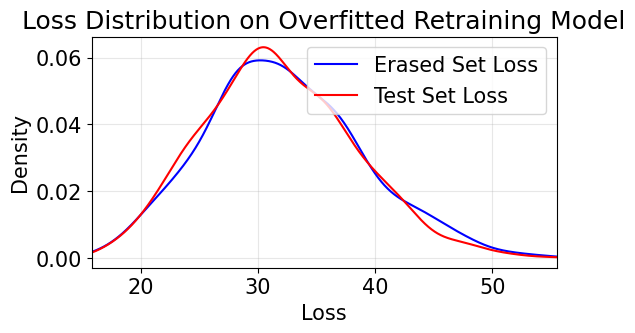}
    \caption{Loss distributions for samples on non-overfitted and overfitted, original and retraining models.}
    \label{fig:loss_distribution}
\end{figure}

\textbf{Test Loss Distributions.} In our MIA experiments, $D_e$ and $D_{test}$ samples are drawn from the same distribution, e.g., samples of the same class. This results in similar loss distributions on $D_e$ and $D_{test}$ samples on both the original and unlearned models. As shown in Fig.~\ref{fig:loss_distribution}, both non-overfitted and overfitted models result in similar loss distributions on $D_e$ and $D_{test}$ samples. This makes it more challenging for the MIA attacker to perform the binary classification, which thus results in not very high accuracy for both original and unlearned models. 

\begin{table*}[t]
    \centering
    \resizebox{0.72\textwidth}{!}{
    \begin{tabular}{c|ccccc}
        \toprule
         \multirow{2}{*}{$\mathcal{M}$} & \multicolumn{5}{c}{\textbf{ResNet18 $\times$ CIFAR-10 Class Removal}}  \\
         \cmidrule(l){2-6}  
         & $D_e$ Acc. ($\rightarrow$) & $D_r$ Acc. ($\uparrow$) & $D_{test}$ Acc. ($\uparrow$) & JS Div. ($\downarrow$) & MIA Acc. \\ \midrule
         Retraining (reference)  & 0.0$\pm$0.0  & 85.85$\pm$1.45    & 75.16$\pm$2.01   & 0.0$\pm$0.0 & 51.30$\pm$2.33  \\ \midrule
         Original  & 96.55$\pm$0.69    & 84.62$\pm$0.12  & 82.77$\pm$0.28   & 0.027$\pm$0.004 & 48.73$\pm$2.10 \\
         Rand. Lbls. & \textbf{0.0}$\pm$0.0  & 18.72$\pm$2.08   & 18.07$\pm$2.24   & \textbf{0.011}$\pm$0.001 & 50.87$\pm$0.59 \\
         GD      & 90.13$\pm$11.18    & \textbf{89.65}$\pm$1.54   & \textbf{87.06}$\pm$0.47   & 0.020$\pm$0.008 & 49.33$\pm$0.17 \\
         GA     & \underline{5.55}$\pm$9.62      & 25.25$\pm$1.78   & 22.64$\pm$2.36   &  0.028$\pm$0.003 &  50.00$\pm$0.0 \\
         SCRUB  & \textbf{0.0}$\pm$0.0	& \underline{88.42}$\pm$0.08	& \underline{77.52}$\pm$0.22 & \underline{0.019}$\pm$0.012 & 50.30$\pm$1.15  \\
         \midrule	
         SCureNewton (ours) & 17.01$\pm$9.70 & 84.06$\pm$2.33  & 75.27$\pm$2.20  & 0.020$\pm$0.017 & 50.07$\pm$0.65 \\
         \bottomrule
    \end{tabular}
    }
    \caption{Performance and MIA results of SCureNewton and other tested baselines in sequential unlearning setting on ResNet18 $\times$ CIFAR-10 (for $3$ random runs). Results are reported at the last unlearning round.
    }
    \label{table:sequential_unlearning_MIA}
\end{table*}

\begin{table*}[t]
    \centering
    \resizebox{0.72\textwidth}{!}{
    \begin{tabular}{c|ccccc}
        \toprule
         \multirow{2}{*}{$\mathcal{M}$} & \multicolumn{5}{c}{\textbf{Overfitted ResNet18 $\times$ CIFAR-10 Class Removal}} \\ \cmidrule(l){2-6}  
         & $D_e$ Acc. ($\rightarrow$) & $D_r$ Acc. ($\uparrow$) & $D_{test}$ Acc. ($\uparrow$) & JS Div. ($\downarrow$) & MIA Acc.  \\
         \midrule
         Retraining (reference) & 0.00$\pm$0.00 & 99.94$\pm$0.01 & 84.52$\pm$0.23 & 0.0$\pm$0.0 & 51.77$\pm$1.59 \\
         \midrule
         Original & 99.89$\pm$0.03 & 99.90$\pm$0.01 & 92.73$\pm$0.07 & 0.036$\pm$0.0  & 58.37$\pm$0.74  \\
         Rand. Lbls. & 16.45$\pm$1.92 & \underline{94.96}$\pm$0.64 & \underline{81.25}$\pm$1.07 & 0.024$\pm$0.0 & 58.47$\pm$0.26  \\
         GD & 99.97$\pm$0.03 & \textbf{99.86}$\pm$0.01 & \textbf{92.53}$\pm$0.17 & 0.036$\pm$0.0 & 57.17$\pm$0.29  \\
         GA & 4.35$\pm$1.36 & 43.64$\pm$14.91 & 37.10$\pm$12.14 & 0.029$\pm$0.001  & 51.17$\pm$1.25 \\
         SCRUB & \textbf{0.00}$\pm$0.00 & 85.92$\pm$1.33 & 75.43$\pm$0.83 & \textbf{0.008}$\pm$0.002 & 50.10$\pm$1.15  \\
         \midrule
         SCureNewton (Ours) & \underline{1.82}$\pm$1.97 & 87.25$\pm$0.77 & 74.52$\pm$0.67 & \underline{0.017}$\pm$0.003  & 54.03$\pm$1.10  \\
        \bottomrule
    \end{tabular}
    }
    \caption{Performance and MIA results for SCureNewton and other tested baselines in sequential unlearning setting on the overfitted ResNet18 $\times$ CIFAR-10 (for $3$ random runs). Results are reported at the last unlearning round.
    }
    \label{table:mia_overfit}
\end{table*}

\section{Supplementary Experiments}
\label{app:sup_exp}

    \subsection{Instance-level Sequential Learning}
    \label{app:instance-sequential}
    
    \begin{figure}[t!]
        \centering
        \includegraphics[width=0.6\linewidth]{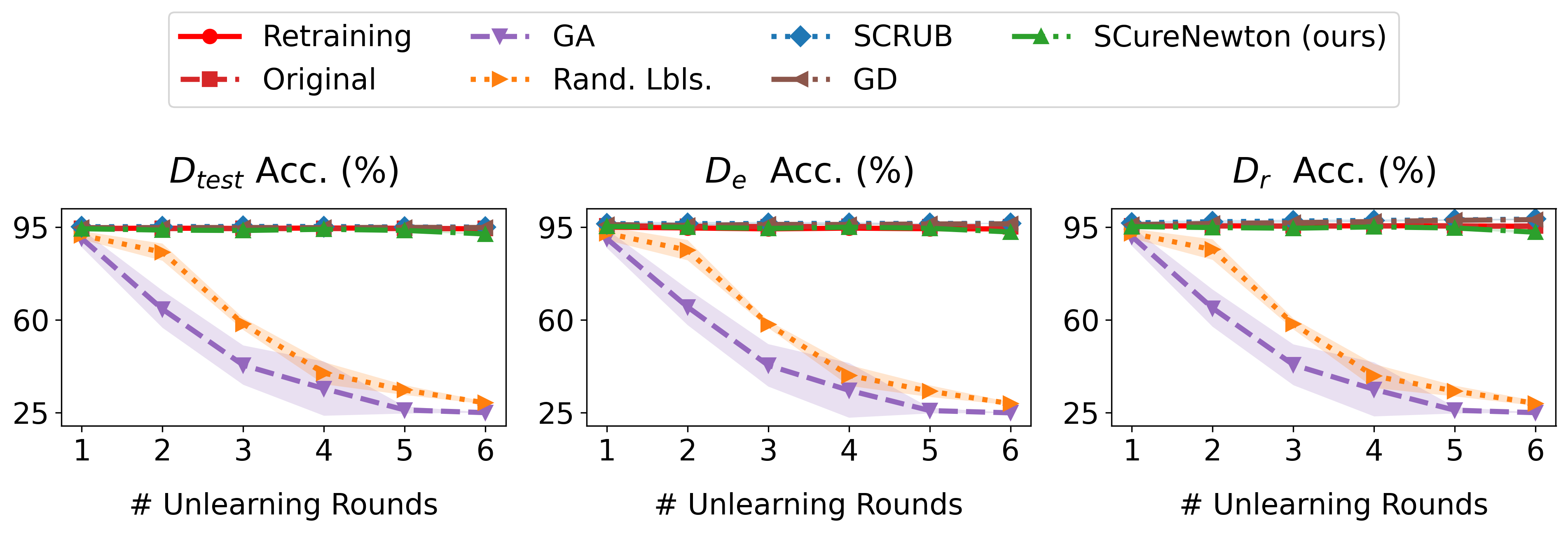} \\
        \includegraphics[width=0.6\linewidth]{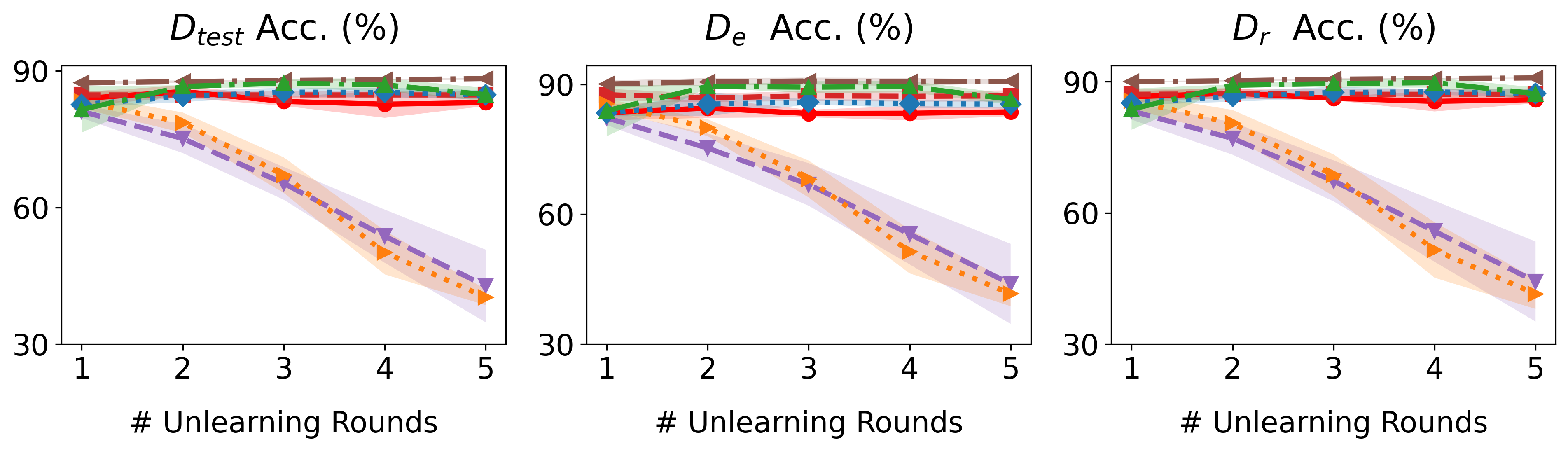}
        \caption{Instance-level sequential unlearning of $10\%$ training data. Top row: Llama-2 $\times$ AG-News ($2000$ erased samples per round). Bottom row: ResNet-18 $\times$ CIFAR-10 ($1000$ erased samples per round).}
        \label{fig:sequential_instance_unlearning}
    \end{figure}
    
    \begin{table*}[t]
        \centering
        \resizebox{0.95\textwidth}{!}{
        \begin{tabular}{c|cccccccc}
            \toprule
             \multirow{2}{*}{$\mathcal{M}$} & \multicolumn{4}{c}{\textbf{AG-News}} & \multicolumn{4}{c}{\textbf{CIFAR-10}} \\
             \cmidrule(l){2-5} \cmidrule(l){6-9} 
             & $D_e$ Acc. ($\rightarrow$) & $D_r$ Acc. ($\uparrow$) & $D_{test}$ Acc. ($\uparrow$) & JS Div. ($\downarrow$) & $D_e$ Acc. ($\rightarrow$) & $D_r$ Acc. ($\uparrow$) & $D_{test}$ Acc. ($\uparrow$)  & JS Div. ($\downarrow$) \\ \midrule
             Retraining (reference) & 94.55$\pm$0.26 & 95.41$\pm$0.14 & 94.53$\pm$0.21 & 0.0$\pm$0.0 & 83.64$\pm$0.98  & 85.89$\pm$0.84    & 83.01$\pm$0.76 &  0.0$\pm$0.0 \\ \midrule
             Original   & \textbf{95.36}$\pm$0.13    & 95.50$\pm$0.01    & 94.63$\pm$0.08 & 0.027$\pm$0.013 & 87.31$\pm$1.29    & 87.11$\pm$1.13    & 84.74$\pm$0.81    & 0.028$\pm$0.004 \\
             Rand. Lbls. & 28.17$\pm$0.57     & 28.21$\pm$0.86   & 28.23$\pm$0.88 & \textbf{0.014}$\pm$0.011 & 41.61$\pm$2.87      & 41.46$\pm$3.43   & 40.28$\pm$1.74   & 0.013$\pm$0.0 \\
             GD         & 96.39$\pm$0.13    & \underline{97.90}$\pm$0.12    & \underline{94.84}$\pm$0.16 & 0.021$\pm$0.015 & 90.72$\pm$0.37    & \textbf{90.86}$\pm$0.10    & \textbf{88.30}$\pm$0.23    & \textbf{0.002}$\pm$0.0 \\
             GA         & 24.91$\pm$0.09    & 25.02$\pm$0.01   & 25.02$\pm$0.02  & \underline{0.018}$\pm$0.002 & 43.90$\pm$9.26      & 44.29$\pm$9.16   & 42.75$\pm$7.98 & 0.032$\pm$0.0 \\
             SCRUB & 96.46$\pm$0.21 & \textbf{98.00}$\pm$0.29 & \textbf{95.11}$\pm$0.11 & 0.028$\pm$0.010 & \textbf{85.97}$\pm$0.78 & \underline{87.63}$\pm$0.51 & \underline{84.85}$\pm$0.64 & \underline{0.003}$\pm$0.0\\ 
             \midrule
             SCureNewton (ours) & \underline{93.73}$\pm$0.87       & 93.71$\pm$1.07    & 92.93$\pm$0.84 & \underline{0.018}$\pm$0.012 & \underline{86.54}$\pm$1.24 & 87.25$\pm$1.44  & 84.77$\pm$1.44    & \underline{0.003}$\pm$0.0 \\
             \bottomrule
        \end{tabular}
        }
        \caption{Performance comparison between SCureNewton and other tested baselines in instance-level sequential unlearning settings on Llama-2 $\times$ AG-News and ResNet18 $\times$ CIFAR-10. 
        Results are reported at the last unlearning round.
        }
        \label{table:instance_sequential_unlearning}
    \end{table*}

    In this experiment, we perform sequential unlearning to iteratively remove a random training subset across multiple rounds, which we refer to as {\it instance-level sequential unlearning}. This experiment aims to complement the {\it class-level sequential unlearning} in Sec. 6.4, where we perform multiple unlearning rounds such that the entire class is removed at the last round.
    Fig.~\ref{fig:sequential_instance_unlearning} and Table~\ref{table:instance_sequential_unlearning} show our results for instance-level sequential unlearning of $10\%$ randomly selected training data on CIFAR-10 and AG-News.
    As can be observed, SCureNewton maintains a close performance to Retraining on $D_e$ and does not degrade model performance on $D_{test}$ and $D_r$ even after multiple unlearning requests. This reinforces our argument that SCureNewton is a good unlearning algorithm for long-term settings such as sequential unlearning on both class and instance levels.

\subsection{Efficiency of Unlearning}
\label{app:exp:efficiency}

    \begin{table}[t]
        \centering
        \resizebox{0.7\linewidth}{!}{
        \begin{tabular}{c|cc|cc|cc}
            \toprule
              Dataset & \multicolumn{2}{c}{\bf FashionMNIST} & \multicolumn{2}{c}{\bf AG-News} & \multicolumn{2}{c}{\bf CIFAR-10} \\ \midrule
              Model & \multicolumn{2}{c}{2-layer CNN} & \multicolumn{2}{c}{Llama-2-7B (+LoRA)} & \multicolumn{2}{c}{ResNet18}  \\ \midrule
             \# Model Parameters & \multicolumn{2}{c}{20,728} & \multicolumn{2}{c}{1,064,960} & \multicolumn{2}{c}{11,173,962}  \\ 
             \midrule
             Retraining & 61.20$\pm$8.70 & 1.0$\times$ & 4792.44$\pm$145.90 $\quad$ & 1.0$\times$  & 124.51$\pm$10.95 & 1.0$\times$  \\
             Rand. Lbls. & 1.70$\pm$0.19 & 0.03$\times$ & 144.63$\pm$1.83 & 0.03$\times$ & 2.58$\pm$0.10 & 0.02$\times$ \\
             GD & 9.04$\pm$0.82 & 0.1$\times$ & 4641.50$\pm$407.93 & 0.96$\times$ & 19.16$\pm$4.03 & 0.2$\times$ \\
             GA & 2.28$\pm$0.58 & 0.03$\times$ & 105.46$\pm$2.85 & 0.02$\times$ & 5.78$\pm$0.26 & 0.04$\times$ \\
             NTK & 676.24$\pm$34.60 & 11.0$\times$ & NA & NA & NA & NA \\
             PINV-Newton & 6185.72$\pm$804.94 & 101.1$\times$ & NA & NA & NA & NA \\
             Damped-Newton & 6228.82$\pm$739.82 & 101.7$\times$ & NA & NA & NA & NA \\
             SCRUB & \textbf{23.33}$\pm$0.43 & 0.4$\times$ & \underline{6796.16}$\pm$160.11 & 1.4$\times$ & \underline{72.39}$\pm$4.93 & 0.6$\times$  \\
             CureNewton (ours) & 6355.31$\pm$127.31  & 103.8$\times$ & NA & NA & NA & NA \\ 
             SCureNewton (ours) & \underline{35.54}$\pm$6.73 & 0.6$\times$ & \textbf{85.26}$\pm$18.23 & 0.02$\times$ & \textbf{41.79}$\pm$0.94 & 0.3$\times$ \\
             \bottomrule
        \end{tabular}
        }
        \caption{Running time comparison (in seconds) across different datasets and models (from $3$ random runs)}. 

        \label{tab:efficiency}
        \vspace{-3mm}
    \end{table}
    
    Table~\ref{tab:efficiency} (an expanded version of Table 3 in the main paper) shows the running time comparison (in seconds) among different unlearning algorithms to unlearn a batch of erased data points across various datasets and models. As anticipated, the unlearning algorithms that utilize the second-order information, such as NTK, PINV-Newton, Damped-Newton, and CureNewton, have the longest running times and even exceed that of Retraining on FashionMNIST. Therefore, these algorithms are impractical for large-scale experiments with CIFAR-10 and AG-News datasets. On the other hand, GA, GD, and Rand. Lbls. are fast unlearning algorithms but tend to significantly degrade model performance post-unlearning, especially in long-term settings such as sequential unlearning (Sections 6.4 and \ref{app:instance-sequential}).
    In contrast, SCureNewton can maintain efficiency across various datasets and models and be more efficient than the state-of-the-art method SCRUB despite being a second-order method by leveraging the fast Hessian-vector products and the stochastic setup; while maintaining a decent erasing quality and post-unlearning performance.

\section{Supplementary Ablation}
\label{app:sup_abl}

    \subsection{Sensitivity of $L$ and $M$ Hyperparameters}

    \begin{figure}
        \centering
        \includegraphics[width=0.40\linewidth]{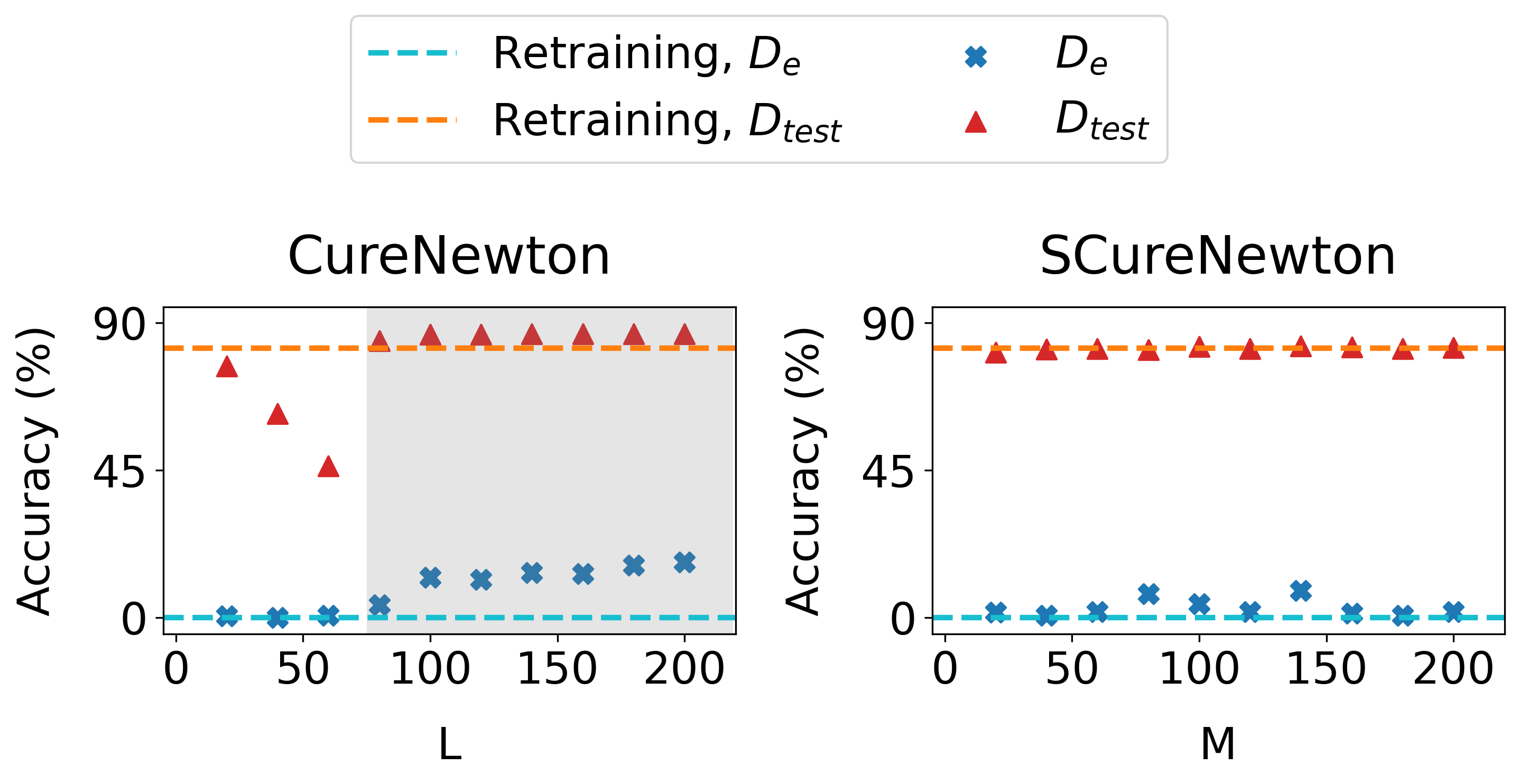}
        \caption{Sensitivity to $L$ and $M$ hyperparameters in CureNewton's and SCureNewton method.}
        \label{fig:L_sensitivity}
    \end{figure}

    Fig.~\ref{fig:L_sensitivity} shows the unlearning performance of CureNewton and SCureNewton with different values of  Hessian Lipschitz $L$ or stochastic Hessian Lipschitz $M$ in sequential unlearning settings on FashionMNIST. As can be seen, both algorithms achieve consistent results and remain close to Retraining across different $L$ or $M$ values. This implies the robustness of our proposed methods in terms of hyperparameters $L$ and $M$, which is contrary to the dependence on the step size in the first-order algorithms (e.g., GD, GA, Rand. Lbls.).
    On a separate note, it is crucial to keep $L$ above a certain threshold (i.e., approximately $L=70$ for CureNewton, as marked by the shaded region, considering $L$ denotes the maximal difference in the Hessians between any two sets of parameters. Yet, it is worth noting that $L$ can be properly tuned/determined in the training phase and before unlearning happens. Therefore, we believe it is appealing to use CureNewton and SCureNewton in practical settings without much hyperparameter tuning.

\subsection{Number of Stochastic Iterations in SCureNewton's Method}
\label{app:stochastic:steps}

\begin{figure}[t]
\centering
     \includegraphics[width=0.65\linewidth]{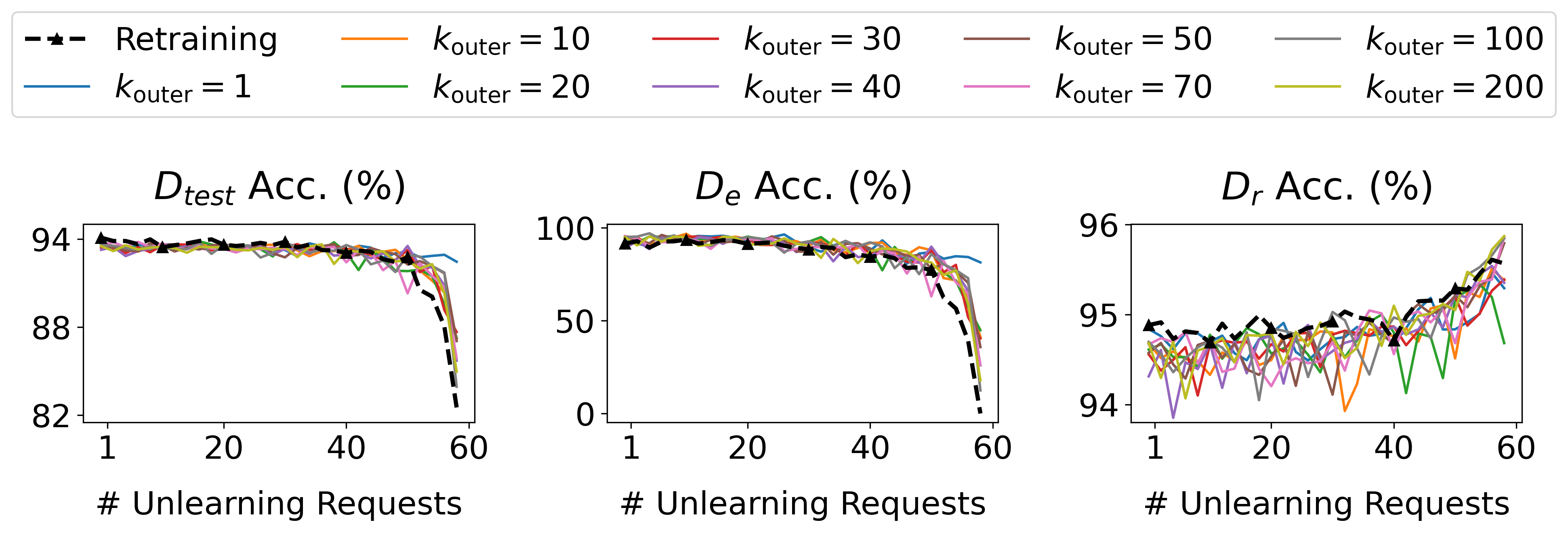}
    \caption{SCureNewton's performance for different $k_\text{outer}$ stochastic iterations in sequential unlearning settings on FashionMNIST.
    }
    \label{fig: stochastic step}
\end{figure}

Fig.~\ref{fig: stochastic step} shows the unlearning performance of SCureNewton's method for different $k_{\text{outer}}$ stochastic iterations in sequential unlearning settings on FashionMNIST. 
As can be seen, the performance on $D_{test}$, $D_e$ and $D_r$ becomes closer to Retraining when a larger number of iterations are adopted. 
This indicates a trade-off between the unlearning performance and time efficiency in SCureNewton.
Consequently, we believe that the number of stochastic iterations in SCureNewton's method is subject to change in order to fulfill specific requirements in practice, where increasing the number of iterations typically improves unlearning performance but may incur longer computational time.

\section{Related Work: Cubic Regularization in Optimization Context}

As described in Sec. 3.2 and 5.1, unlearning can be approached from the perspective of learning/optimization on the retained data points (i.e. excluding the data points to be erased). Therefore, we believe it is useful to understand how cubic regularization fits within the optimization context, as this can help us draw connections to the unlearning algorithms proposed in our paper.

{\bf Convex optimization.} Among optimization algorithms, gradient descent is probably the most commonly used one in the machine learning community due to its computational efficiency, especially on deep neural networks.
For a convex loss function $f$ with $\ell$-gradient Lipschitz continuity, it has been proven that gradient descent can find $\epsilon$-first-order stationary points\footnote{An $\epsilon$-first-order stationary point satisfies $\lVert \nabla f(x) \rVert \leq \epsilon$.}  in $\ell(f(x_0) - f(x^*))/\epsilon^2$ iterations, where $x_0$ is the initial point and $x^*$ is the value that optimizes $f$ \cite{nesterov13:introductory}. Since the Hessian matrix is positive semidefinite for convex losses, the first-order stationary point is also a second-order stationary point and a global optimum of the loss function.

{\bf Non-convex optimization.}
Many real-world functions, such as neural network losses, are non-convex.
Non-convex optimization is more challenging since the landscape of the non-convex functions often contains many saddle points and their local minima may differ from the global minima.
Therefore, convergence to a first-order stationary point (as in gradient descent) is not sufficient, as the point can be either a local minimum, a local maximum, or a saddle point. 
Concurrently, previous works have shown that local minima can be as effective as global minima in many situations \cite{sun15:nonconvex-not-scary, ge16:matrix-local}.
This observation has motivated the development of optimization algorithms that converge to $\epsilon$-second-order stationary points~\footnote{An $\epsilon$-second-order stationary point satisfies $\lVert \nabla f(x) \rVert \leq \epsilon$ and $\lambda_{min}(\nabla^2 f(x)) \geq -\epsilon$.} or local minima.

Newton's method is one of the most popular second-order methods that leverages the curvature information of the function and achieves a local convergence to the $\epsilon$-second-order stationary point in $O(\frac{1}{\epsilon^2})$ iterations; while its global convergence is often known to be much slower.
Extended from Newton's method, \citeauthor{nestero06:cubic-newton} proposes cubic regularized Newton's method, which improves global convergence to the $\epsilon$-second-order stationary points in $O(\frac{1}{\epsilon^{1.5}})$. However, cubic regularized Newton's method needs access to the full Hessian and full gradient over the entire dataset,  which can be costly in terms of both computation time and storage.
To address this problem, many works have adapted cubic regularized Newton's method to stochastic settings. 
\citet{kohler17:subsample} uses subsampled Hessian and gradient but does not provide asymptotic analysis for their algorithms. 
\citet{xu20:inexact-hessian} considers stochastic Hessians but still requires gradients on the entire dataset.
\citet{tripuraneni18:stochastic-cubic} can efficiently employ stochastic gradient and stochastic Hessian-vector products while converging to $\epsilon$-second-order stationary points in $\Tilde{O}(\frac{1}{\epsilon^{3.5}})$ iterations where $\Tilde{O}$ hides the poly-logarithms factor.

\end{document}